\documentclass[pdflatex,sn-mathphys-num,iicol]{sn-jnl}

\usepackage{graphicx}%
\usepackage{multirow}%
\usepackage{amsmath,amssymb,amsfonts}%
\usepackage{amsthm}%
\usepackage{mathrsfs}%
\usepackage[title]{appendix}%
\usepackage{xcolor}%
\usepackage{textcomp}%
\usepackage{manyfoot}%
\usepackage{booktabs}%
\usepackage{algorithm}%
\usepackage{algorithmicx}%
\usepackage{algpseudocode}%
\usepackage{listings}%
\usepackage{algpseudocode}
\usepackage{tabularx}
\usepackage{utfsym}
\usepackage{float}
\usepackage[algo2e,ruled,vlined]{algorithm2e}

\usepackage{booktabs}
\usepackage{mathtools}
\usepackage{caption} 
\usepackage{makecell}

\usepackage[dvipsnames]{xcolor}

\SetCommentSty{mycommfont}

\theoremstyle{thmstyleone}%
\theoremstyle{thmstyletwo}%

\theoremstyle{thmstylethree}%

\begin{document}

\title[Article Title]{Knowledge is Power:  Advancing Few-shot Action Recognition with Multimodal Semantics from MLLMs}


\author[1]{\fnm{Jiazheng} \sur{Xing}}\email{jiazhengxing@zju.edu.cn}

\author[1]{\fnm{Chao} \sur{Xu}}\email{21832066@zju.edu.cn}

\author[1]{\fnm{Hangjie} \sur{Yuan}}\email{hj.yuan@zju.edu.cn}

\author[2]{\fnm{Mengmeng} \sur{Wang}}\email{wangmengmeng@zjut.edu.cn}

\author[3]{\fnm{Jun} \sur{Dan}}\email{danjun@zju.edu.cn}

\author[4]{\fnm{Hangwei} \sur{Qian}}\email{qian\_hangwei@a-star.edu.sg}

\author*[1]{\fnm{Yong} \sur{Liu}}\email{yongliu@iipc.zju.edu.cn}

\affil*[1]{\orgdiv{College of Control Science and Engineering}, \orgname{Zhejiang University}, \orgaddress{\city{Hangzhou}, \postcode{310027}, \state{Zhejiang}, \country{China}}}

\affil[2]{\orgdiv{College of Computer Science and Technology}, \orgname{Zhejiang University of Technology}, \orgaddress{\city{Hangzhou}, \postcode{310023}, \state{Zhejiang}, \country{China}}}

\affil[3]{\orgdiv{College of Information Science and Electronic Engineering}, \orgname{Zhejiang University}, \orgaddress{\city{Hangzhou}, \postcode{310027}, \state{Zhejiang}, \country{China}}}

\affil[4]{\orgdiv{CFAR and IHPC}, \orgname{A*STAR}, \orgaddress{\postcode{138632}, \country{Singapore}}}


\abstract{Multimodal Large Language Models (MLLMs) have {propelled} the field of few-shot action recognition (FSAR). However, preliminary explorations in this area focus primarily on generating captions to form a suboptimal feature$\rightarrow$caption$\rightarrow$feature pipeline and adopt metric learning solely within the visual space. In this paper, we propose \textbf{FSAR-LLaVA}, the first end‑to‑end method to leverage MLLMs (such as Video‑LLaVA) as a multimodal knowledge base for directly enhancing FSAR. First, at the feature level, we leverage the MLLM’s multimodal decoder to extract spatiotemporally and semantically enriched representations, which are then decoupled and enhanced by our {Multimodal Feature-Enhanced Module} into distinct visual and textual features that fully exploit their semantic knowledge for FSAR. Next, we leverage the versatility of MLLMs to craft input prompts that flexibly adapt to diverse scenarios, and use their aligned outputs to drive our designed Composite Task‑Oriented Prototype Construction, effectively bridging the distribution gap between meta‑train and meta‑test sets. Finally, to enable multimodal features to guide metric learning jointly, we introduce a training-free Multimodal Prototype Matching Metric that adaptively selects the most decisive cues and efficiently leverages the decoupled feature representations produced by MLLMs. Extensive experiments demonstrate superior performance across various tasks with minimal trainable parameters.}


\maketitle

\section{Introduction}
Few-shot action recognition (FSAR) seeks to rapidly learn a generalized model capable of classifying unseen action categories using limited labeled samples. Unlike action recognition (AR)~\cite{wang2016temporal,wang2022learning, wang2023actionclip,gao2020pairwise}, which demands considerable resources for data collection and annotation, FSAR holds greater promise for industrial applications due to its distinctive characteristics. To tackle the challenge of recognizing new classes in FSAR with limited sample information, recent efforts~\cite{cao2020few,perrett2021temporal,wang2022hybrid, xing2023boosting} mainly focus on the metric-learning paradigm to learn to construct class prototypes through episode training and employ a robust alignment metric for matching. 

\begin{figure} [t!]
\centering
\includegraphics[width=\linewidth]{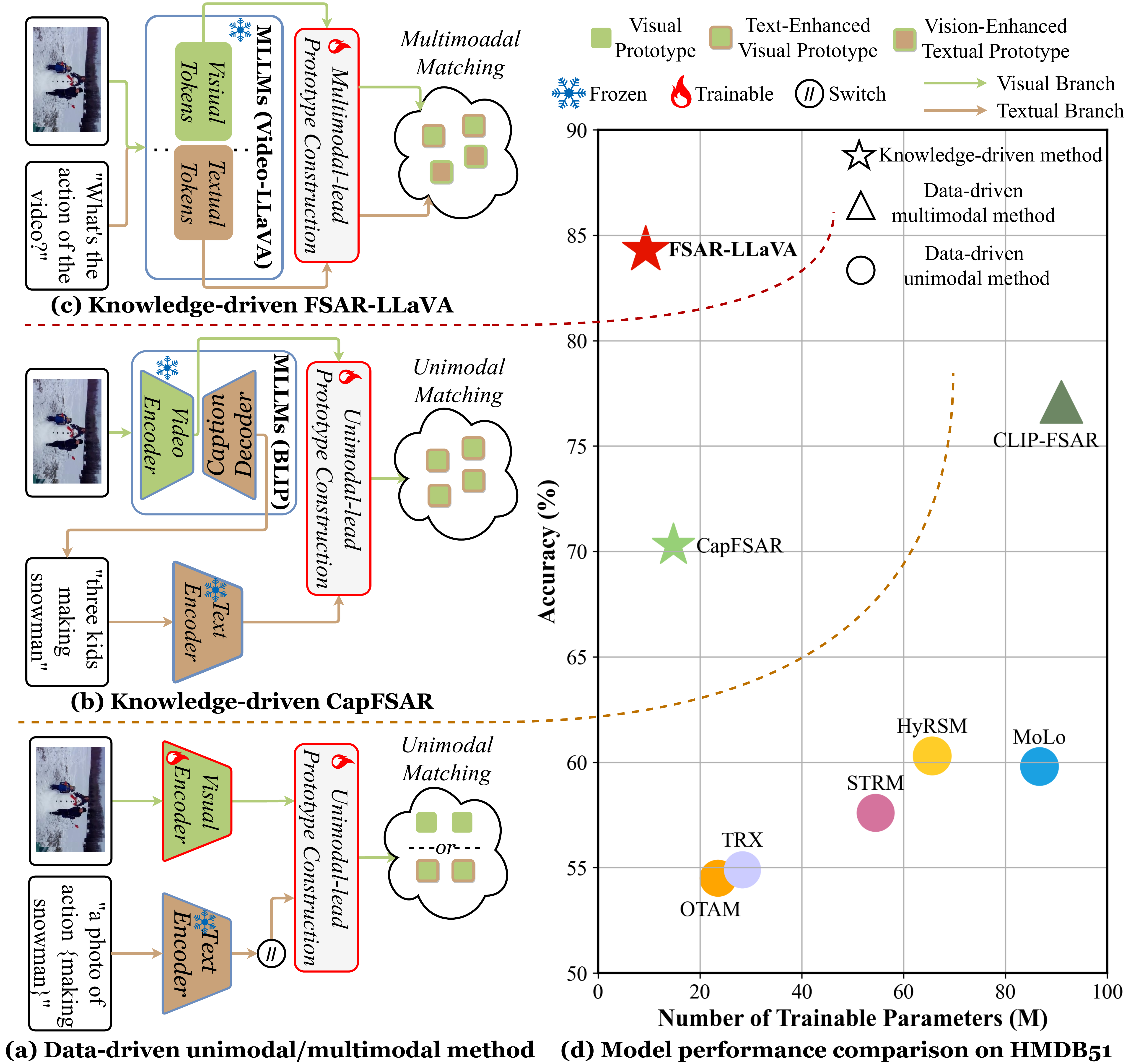}
\caption{
(a), (b), and (c) indicate the pipeline of the data-driven unimodal/multimodal method, knowledge-driven CapFSAR, and our FSAR-LLaVA, respectively. 
Our {FSAR-LLaVA$_\mathrm{Unknown}$}, which uses the fixed input prompt:   ``What's the action of the video?" without introducing additional textual label information, fully leverages the multimodal features of MLLM and achieves state‑of‑the‑art performance that requires minimal parameters, as depicted in part (d), which refers to the performance comparison in the HMDB51 5-way 1-shot task.}
\label{fig:ab}
\end{figure}

Based on the metric-learning paradigm, the previous dominant trend is data-driven FSAR, which can be divided into two categories: unimodal methods and multimodal methods, as shown in Fig.~\ref{fig:ab}(a). 
Unimodal methods~\cite{cao2020few,wang2022hybrid} rely solely on visual information from the task-provided videos.
However, they struggle to distinguish similar categories of actions in low-shot settings due to the limited number of support-set videos.
To address this, the follow-up research efforts~\cite{wang2023clip,qu2024mvp, xing2026ma} introduce text semantic information into FSAR to guide the construction of class prototypes by incorporating the manually labeled texts with the help of discriminative multimodal models, {such as} CLIP~\cite{radford2021learning}. 
However, manually annotating text is laborious and requires expertise.
CapFSAR~\cite{wang2023few} takes the first step to caption videos by utilizing the multimodal priors of Multimodal Large Language Models (MLLMs), thereby eliminating the need for manual text annotation, as shown in Fig.~\ref{fig:ab}(b).
Nevertheless, its pipeline {that decodes features into captions and then re-encodes them to features (feature$\rightarrow$caption$\rightarrow$feature)} significantly {undermines} the efficacy of utilizing multimodal knowledge due to \textit{information degradation in the intermediate captioning step}, where visual features are translated into natural language and then remapped back into the latent space.
Moreover, the metric learning solely within the visual space neglects the multimodal nature of the MLLM knowledge base.

Recent research has revealed that MLLMs possess a wealth of {real-world} knowledge and thus can be integrated into downstream tasks to inherit their superior abilities in multimodal understanding and reasoning. 
Research efforts, spanning visual detection~\cite{zang2023contextual}, grounding~\cite{peng2023kosmos}, and segmentation~\cite{wang2024llm}, have resorted to MLLMs for direct and effective knowledge acquisition, such as BLIP~\cite{li2022blip} and LLaVA~\cite{liu2024visual}.
Inspired by these attempts, a natural question arises: 
{\emph{Can MLLMs' vast world knowledge be seamlessly and directly integrated into FSAR without the information degradation introduced by intermediate steps?}}

{In this paper, we introduce FSAR‑LLaVA, the first end‑to‑end, knowledge‑driven FSAR framework that harnesses the extensive knowledge of MLLMs effectively, with no information loss.}
Firstly, we process the multimodal features from the multimodal decoder of MLLMs (\textit{e.g.}, Video-LLaVA) and comprehensively explore their spatiotemporal perception and semantic understanding capabilities across different layers. 
{Secondly, we decouple the multimodal features into distinct text and vision branches and design a Multimodal Feature-Enhanced Module with a symmetric dual-branch structure that interactively refines each modality.} 
Thirdly, we further explore the properties of MLLMs. 
For MLLMs' outputs in FSAR, we observe that similar prompt inputs generate answers in a consistent pattern. This aligned characteristic inspires us to design {local prototype construction}, which combined with the common {global prototype construction}, forms our proposed Composite Task-Oriented Prototype Construction. 
{This composite design dynamically refines support‑set class centers by leveraging the strengths of MLLMs, effectively closing the distribution gap between meta‑train and meta‑test sets.}
Finally, coupled with a tailored Multimodal Prototype Matching Metric, we can determine which elements play a decisive role during the matching phase, thus flexibly and sufficiently leveraging multimodal features.
All of the above modules involve only a minimal number of training parameters for high efficiency. 
In summary, our knowledge-driven FSAR-LLaVA fully leverages the world knowledge extracted from MLLMs, achieving an optimal balance between performance and computational cost compared to other approaches, as shown in Fig.~\ref{fig:ab}(d). 
The contributions can be summarized as follows:
\begin{itemize}
    \item {We propose FSAR‑LLaVA, the first end‑to‑end, knowledge‑driven FSAR framework that incorporates real‑world knowledge extracted from MLLMs, thereby reducing the dependency on high‑quality data.}
    \item 
    {To efficiently adapt MLLMs across the entire FSAR pipeline}, we meticulously develop a Multimodal Feature-enhanced Module, a Composite Task-oriented Prototype Construction Module, and a Multimodal Matching Metric, {thereby fully harnessing MLLMs’ capabilities to substantially enhance FSAR performance.}
    \item Extensive experiments unequivocally demonstrate that our method attains exceptional performance while employing the fewest trainable parameters.
\end{itemize}

\section{Related Work}

\subsection{Multimodal Large Language Models (MLLMs)}
Following the introduction of the widely recognized commercial model ChatGPT~\cite{achiam2023gpt}, the AI community is actively responding by developing and releasing open-source Large Language Models (LLMs), including LLaMA~\cite{touvron2023llama}, Vicuna~\cite{chiang2023vicuna}, Qwen2~\cite{yang2024qwen2}, Qwen3~\cite{yang2025qwen3technicalreport} and others. LLMs can be expanded into Multimodal Large Language Models (MLLMs) when visual sources participate.  
Mainstream MLLMs such as BLIP~\cite{li2022blip}, MiniGPT-4~\cite{zhu2023minigpt}, and LLaVA~\cite{liu2024improved} typically treat the LLM as a decoder for multimodal understanding.
For {video understanding}, MLLMs such as VideoChat~\cite{li2023videochat}, Video-LLaMA~\cite{zhang2023video},
Video-LLaVA~\cite{lin2023video}, Qwen2-VL~\cite{wang2024qwen2}, and Qwen3-VL~\cite{bai2025qwen3vltechnicalreport}, which handle videos, also need to focus on temporal modeling. {Beyond text generation, prior work shows that the decoder-only LLM’s hidden state of a special token can be used as a vision prompt/representation and then decoded for dense prediction. For example,
LISA~\cite{lai2024lisa} adds a \texttt{<SEG>} token and decodes its hidden embedding into a segmentation mask,
and VideoLISA~\cite{bai2024one} extends this to videos with a \texttt{<TRK>} token and uses its last hidden embedding to prompt the mask decoder for temporally consistent segmentation/tracking across frames.}
In this work, we use Video-LLaVA as our primary knowledge base and also conduct additional experiments on Qwen2-VL and Qwen3-VL to evaluate our framework's generalization.
{Unlike LISA/VideoLISA, which introduce special tokens and decode their hidden states via a mask decoder to obtain dense segmentation masks, our FSAR-LLaVA focuses on {few-shot action recognition} and directly leverages intermediate
multimodal hidden features from video MLLM decoders as transferable representations.}

\subsection{Few-shot Action Recognition (FSAR)}
Research on few-shot learning (FSL) is primarily categorized into adaptation-based and metric-based methods. Adaptation-based methods~\cite{finn2017model,nichol2018reptile} focus on identifying a network initialization that can be fine-tuned for novel tasks with limited labeled data, a process known as \textit{gradient by gradient}.  Metric-based methods~\cite{snell2017prototypical,vinyals2016matching} aim to learn 
a feature mapping space and compare task features using different matching strategies, a concept referred to as \textit{learning to compare}. Most of the current mainstream few-shot action recognition (FSAR) methods follow the metric-based manner driven by limited data in each episode to optimize the model and design alignment metrics to calculate the distances between the query and support samples for recognition. These data-driven methods can be classified into two categories: 1)  unimodal, which only uses the original video sources, and 2) multimodal, which incorporates both the video sources and the text priors from the support samples. For data-driven unimodal methods, some focus on the exploration of matching metrics. OTAM~\cite{cao2020few} introduces a temporal alignment metric, TRX~\cite{perrett2021temporal} matches each query sub-sequence with all sub-sequences in the support set, and HyRSM~\cite{wang2022hybrid} proposes a bidirectional Mean Hausdorff Metric to robustly align complex actions. Others pay attention to improving features' {or} class prototypes' representation. STRM~\cite{thatipelli2022spatio} proposes local and global enrichment modules for features' spatiotemporal relationship enhancement, HyRSM~\cite{wang2022hybrid} introduces the hybrid relation modeling for obtaining task-specific embeddings, and SloshNet~\cite{xing2023revisiting} designs the short-term and long-term temporal modeling modules. Data-driven multimodal methods~\cite{wang2023clip, qu2024mvp, cao2024task} primarily focus on adapting CLIP~\cite{radford2021learning} to FSAR and leveraging textual priors to effectively guide the semantic understanding of visual features. Recently, a knowledge-driven approach, CapFSAR~\cite{wang2023few}, proposes using BLIP~\cite{li2022blip} to generate text labels for support samples, eliminating the need for manual annotation. However, its exploration of refining MLLMs for FSAR remains insufficient.

\section{Method}

\begin{figure*} [ht!]
	\centering
	\includegraphics[width=\linewidth]{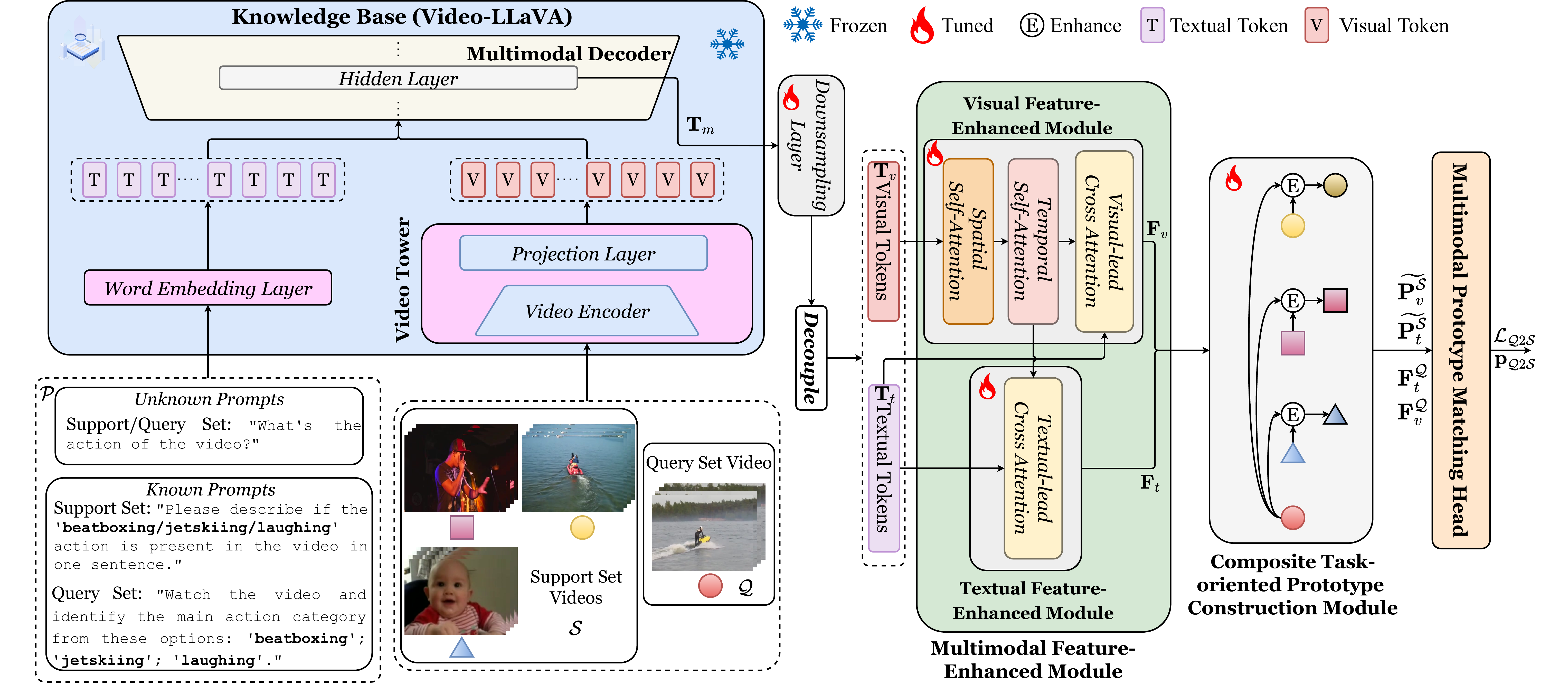}
	\caption{Overview of our  FSAR-LLaVA: Visual inputs and text prompts are processed through our knowledge base to extract multimodal tokens $\textbf{T}_m$ from the multimodal decoder's hidden layer. These tokens are downsampled and decoupled into visual tokens $\textbf{T}_v$ and textual tokens $\textbf{T}_t$. The tokens are then passed through the Multimodal Feature-enhanced Module, and features from both branches are processed in the Composite Task-oriented Prototype Construction Module to obtain the enhanced support set class prototypes $\widetilde{\textbf{P}^\mathcal{S}_v}$ and $\widetilde{\textbf{P}^\mathcal{S}_t}$. Finally, the class prototypes and query features are fed into the Multimodal Matching Metric to yield the probability distribution $\textbf{p}_{\mathcal{Q}2\mathcal{S}}$ and loss  $\mathcal{L}_{\mathcal{Q}2\mathcal{S}}$.
    }
	\label{fig:main}
\end{figure*}

\subsection{Problem Formulation}

{Give a meta-train set $\mathcal{D}_{\mathrm{train}} = \{(x_i, y_i) \mid y_i \in \mathcal{C}_{\mathrm{train}}\}$ and a meta-test set $\mathcal{D}_{\mathrm{test}} = \{(x_j, y_j) \mid y_j \in \mathcal{C}_{\mathrm{test}}\}$, where $y_i$ is the action category of a video sample $x_i$. Note that the categories of $\mathcal{D}_{\mathrm{train}}$ and $\mathcal{D}_{\mathrm{test}}$ are no overlapping, \textit{i.e.} $\mathcal{C}_{\mathrm{train}} \cap \mathcal{C}_{\mathrm{test}} = \varnothing$. The objective of FSAR is to classify an unseen query video using only a few samples during testing, and verify the generalization of the trained model.}
To simulate the few-shot testing environment, we adopt an episode training manner followed by~\cite{zhu2018compound,cao2020few,perrett2021temporal}, selecting episodes randomly from a large pool of collected data. {In each episode, the support set $\mathcal{S}$ contains $N\times K$ videos from $N$ classes with $K$ samples per class, while the query set $\mathcal{Q}$ follows the same $N$-way $K$-shot setting, with a full set of test queries evaluated simultaneously during metric evaluation.} Specifically, for class $n \in\left\{1,\cdots,N\right\}$, the $k$-th video is denoted as $S_k^n=\left\{s_{k1}^n,s_{k2}^n,\cdots,s_{kT}^n\right\}$  containing $T$ frames, while the unseen single query video is represented as $ Q=\left\{q_1,q_2,\cdots,q_T\right\}$.


\subsection{Overview of FSAR-LLaVA}
The overall architecture of our FSAR-LLaVA is shown in Fig.~\ref{fig:main}.  In this pipeline, the visual inputs include the query set videos $\mathcal{Q}$ and the class support set videos $\mathcal{S}$. For frame selection, we follow TSN~\cite{wang2016temporal}, dividing the video into 
$T$ segments and extracting snippets from each segment. For text prompts $\mathcal{P}$, we provide two types: ``Unknown" and ``Known" {(without/with the support set labels)}, to inquire about the action categories of the videos (discussion is shown in Sec.~\ref{discuss_prompt}).  Visual inputs and text prompts pass through our knowledge base (Video-LLaVA~\cite{lin2023video}), leveraging rich real-world knowledge priors to extract discriminative multimodal tokens $\textbf{T}_m$ from the multimodal decoder's \textit{one} hidden layer, which inherently contains implicit pointers to action categories. Next, we downsample the channel dimension of multimodal tokens to reduce computation costs and decouple them into visual tokens $\textbf{T}_v$ and textual tokens $\textbf{T}_t$, thereby accommodating diverse preferences across different datasets (see in Tab. 3). Subsequently, the visual and textual tokens are fed into the lightweight Visual and Textual Feature-Enhanced Module, for stronger feature representation $\textbf{F}_v$ and $\textbf{F}_t$. Afterward, the features from the two branches pass through the Composite Task-oriented Prototype Construction Module to obtain the enhanced support set class prototypes $\widetilde{\textbf{P}^\mathcal{S}_v}$ and $\widetilde{\textbf{P}^\mathcal{S}_t}$. Finally, the dual branches' class prototypes and query features are fed into our tailored Multimodal Matching Metric to obtain the probability distribution $\textbf{p}_{\mathcal{Q}2\mathcal{S}}$ and loss  $\mathcal{L}_{\mathcal{Q}2\mathcal{S}}$.

\subsection{Multimodal Feature-Enhanced Module (MFM)}
{Unlike CapFSAR~\cite{wang2023few}, our method explicitly leverages intermediate multimodal features from MLLMs (serving as the knowledge base) via a dual-branch symmetric MFM that decouples and enhances these features into visual and textual representations, fully exploiting their semantic knowledge for FSAR.}
Given the multimodal tokens $\textbf{T}_m\in \mathbb{R} ^ {B\times L \times D}$ extracted from MLLMs, where $B$ represents the number of videos in one task, and $L$ and $D$ denote the token length and dimension respectively, we apply the Downsampling Layer to reduce computation costs, resulting in $\textbf{T}_m\in \mathbb{R} ^ { B \times L \times D'} (D' < D)$. {To preserve modality-specific semantics and enable effective matching in FSAR}, we decouple the $\textbf{T}_m$ into visual tokens $\textbf{T}_v \in  \mathbb{R} ^ { B \times T \times L_{sp} \times D'}$and textual tokens $\textbf{T}_t \in  \mathbb{R} ^ { B \times L_t \times D'}$, where $T$, $L_{sp}$ indicates the number of frames and the spatial length of each token, and $L = L_t + T \times L_{sp}$. This decoupling is essential to allow modality-specific processing, where visual tokens capture fine-grained appearance/motion patterns and textual tokens convey high-level semantic priors. 

{Specifically, the decoupling method follows the reverse process of multimodal token fusion in Video-LLaVA~\cite{lin2023video}. The complete decoupling process is shown in Algorithm.~\ref{alg:decoup_fsar} of our FSAR-LLaVA.  {\small \texttt{Steps 1}} to {\small \texttt{4}} correspond to the multimodal token fusion process in Video-LLaVA, where the video input $V$, consisting of $T$ frames, is processed.  During this process, the positions of the image tokens, denoted as $\mathcal{I}_{img}$, are obtained. 
Multimodal tokens are passed through the multimodal decoder (LLM) to obtain hidden features at {\small \texttt{Step 5}}. We then decouple them based on the $\mathcal{I}_{img}$, applying the inverse operation of the fusion process as described in {\small \texttt{Step 6}}. Since the multimodal decoder is essentially a transformer, the transformer does not alter each token's original position information or initial properties. Therefore, even after passing through $n$ layers of the transformer, the decoupled tokens can still effectively represent visual and textual information. For utilizing other MLLMs, such as Qwen2-VL~\cite{wang2024qwen2}, as the knowledge base in our framework, its decoupling process remains consistent with that of our FSAR-LLaVA.} 


\begin{algorithm*}[t]
\caption{Decoupling Process in FSAR-LLaVA}
\label{alg:decoup_fsar}
\KwIn{$T$ video frames $V$; Prompt text $\mathcal{P}$:\textit{``What's the action of the video?"}}
\KwOut{Visual tokens $\textbf{T}_v$, Textual tokens $\textbf{T}_t$}

\BlankLine
\textbf{Step1: Model Initialization} \\
$\texttt{(Tokenizer, Model)} \leftarrow \texttt{LoadPretrainedMode}()$ \tcp*{Video-LLaVA Initialization}

\BlankLine
\textbf{Step2: Obtain Input Textual Tokens } \\
$\mathcal{P'} \leftarrow \texttt{Model.BuildPrompt}(\mathcal{P}, T)$ \tcp*{Generate input prompt: ``...\texttt{<}image\texttt{>}...\texttt{<}image\texttt{>} What's...video?..."}
\hspace{0.2em} $I_{\mathcal{P}} \leftarrow \texttt{Tokenizer}(\mathcal{P'})$ \tcp*{Tokenize expanded prompt into IDs}
\hspace{0.2em} $\mathcal{I}_{img} = \{index \ | \ I_{\mathcal{P}}[index]=\texttt{<}image\texttt{>}\}$ \tcp*{Extract positions of image placeholders}
\hspace{0.2em} $\mathbf{T}_{t}^{in} =  \texttt{Model.EmbedTokens}(I_{\mathcal{P}} \  \backslash \hspace{0.2em}   \mathcal{I}_{img})$ \tcp*{Remove image placeholders and embed remaining text}

\BlankLine
\textbf{Step3:  Obtain Input Visual Tokens} \\
$\mathbf{T}_{v}^{in} =  \texttt{Model.VideoEncode}(V)$ \tcp*{Encode video frames into tokens}
\BlankLine

\textbf{Step4:  Tokens Fusion} \\
$\mathbf{T}_{m}^{in} =  \texttt{Model.Fuse}(\mathbf{T}_{t}^{in}, 
\mathbf{T}_{v}^{in}, \mathcal{I}_{img})$ \tcp*{Insert visual tokens into the placeholder positions} \BlankLine

\textbf{Step5:  Hidden Multimodal Features Extraction} \\
$\mathbf{T}_{m} =  \texttt{Model.MultiDecode}(\mathbf{T}_{v}^{in})$ \tcp*{Apply multimodal decoding}
\hspace{0.2em} $\mathbf{T}_{m} =  \texttt{Model.Downsample}(\mathbf{T}_{m})$ \tcp*{Apply downsampling} 
\BlankLine

\textbf{Step6:  Tokens Decoupling} \\
\hspace{0.2em} $\mathbf{T}_{t} =  \texttt{Model.TextualSplit}(\mathbf{T}_{m}, \mathcal{I}_{img})$ \tcp*{Extract textual tokens from multimodal tokens}
\hspace{0.2em} $\mathbf{T}_{v} =  \texttt{Model.VisualSplit}(\mathbf{T}_{m}, \mathcal{I}_{img})$ \tcp*{Extract visual tokens from multimodal tokens}
\end{algorithm*}

For the {decoupled} visual branch enhancement, we introduce the Visual Feature-Enhanced Module. We first perform self-attention followed by global average pooling on the spatial dimension to enhance its representation, obtaining $\textbf{T}_{v\_sp} \in  \mathbb{R} ^ { B \times T \times D'}$. Then, we apply self-attention to its temporal dimension to explore its motion relationships, resulting in $\textbf{T}_{v\_spt} \in  \mathbb{R} ^ { B \times T \times D'}$, as given by:
    \begin{equation}\label{}
    \textbf{T}_{v\_sp} =\mathrm{Avgpool}\left( \mathrm{SelfAttn}(\textbf{T}_v) + \textbf{T}_v\right) 
    \vspace{-8pt}
    \end{equation}
    \vskip -0.08in
    \begin{equation}\label{}
    \textbf{T}_{v\_spt} = \mathrm{SelfAttn}(\textbf{T}_{v\_sp}) + \textbf{T}_{v\_sp}
    \end{equation}
Next, we apply cross-attention to integrate textual information from textual tokens $\textbf{T}_t$ into visual tokens $\textbf{T}_{v\_spt}$ to obtain $\textbf{F}_{v}$, with visual tokens serving as the query, thus enabling a visual-lead fusion, as written by:
 \begin{equation}\label{}
    \textbf{F}_{v} = \mathrm{CrossAttn}(\textbf{T}_{v\_spt}, \textbf{T}_t, \textbf{T}_t) + \textbf{T}_{v\_spt}
    \end{equation}

For the {decoupled} textual branch, we propose the Textual Feature-Enhanced Module. In this module, we apply the textual-lead cross-attention to integrate visual spatiotemporal enhanced tokens $\textbf{T}_{v\_spt}$ into $\textbf{T}_t$ to obtain  $\textbf{F}_{t}$, where textual tokens serving as the query, expressed as: 
 \begin{equation}\label{}
    \textbf{F}_{t} = \mathrm{CrossAttn}( \textbf{T}_t,\textbf{T}_{v\_spt},\textbf{T}_{v\_spt}) +\textbf{T}_t
    \end{equation}

\subsection{Composite Task-oriented Prototype Construction Module (CTPCM)}
\label{CTPCM_intro}
{The feature distributions of the meta-train set {$\mathcal{D}_{train}$} and the meta-test set {$\mathcal{D}_{test}$} are not perfectly aligned—specifically, there are no overlapping categories.} This distributional shift introduces a discrepancy that the prototype constructed by the feature network fails to mitigate. Prior data-driven approaches~\cite{wang2023clip, wang2022hybrid} primarily address this challenge by constructing class prototypes exclusively within the visual branch. 
{By incorporating the knowledge base, we obtain knowledge‑rich visual and textual tokens. In each branch, these tokens leverage their unique strengths to enhance prototype construction, thereby fortifying the framework’s overall robustness.
}

\begin{figure} [t!]
	\centering
	\includegraphics[width=\linewidth]{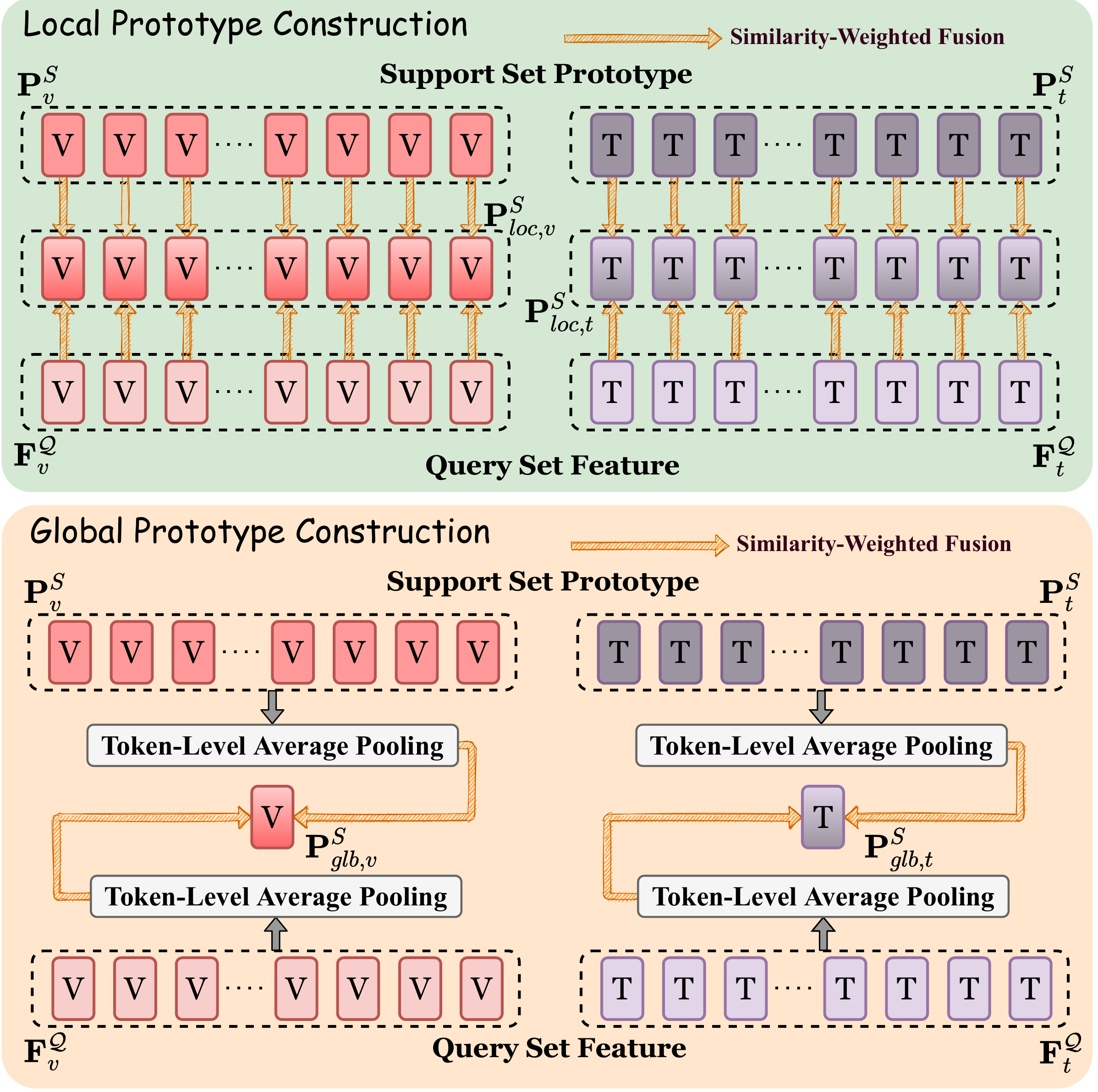}
	\caption{Overview of our CTPCM, which is divided into {local prototype construction} and {{global prototype construction}}.
    }
	\label{fig:CTPCM}
\end{figure}

Specifically, we implement a prototype-level refined design that takes into account the characteristics of Video-LLaVA in this module, leveraging task-oriented query features to adaptively adjust support prototypes, thereby effectively mitigating the distribution gap.
Given the enhanced features $\textbf{F}_v$ and $\textbf{F}_t$ output by MFM, we take the mean value of intra-class support features  $\textbf{F}_v^\mathcal{S}\in \mathbb{R} ^ { 1 \times K \times L_v \times D'}, \textbf{F}_t^\mathcal{S} \in \mathbb{R} ^ { 1 \times K \times L_t \times D'}$ along the shot dimension ($K$) as the support prototype $\textbf{P}^S_i \in \left\{\textbf{P}_v^\mathcal{S}, \textbf{P}_t^\mathcal{S} \right\}$, initially.  
The visual and textual branches independently apply the Composite Task-oriented Prototype Construction Module (CTPCM) within their respective modalities to leverage the task-oriented distribution of query set features, optimizing the support prototype representation in each task.  Specifically, the support prototype and its corresponding branch query set videos are denoted by $\textbf{P}^S_i \in \mathbb{R} ^ { 1 \times L_i \times D'}$   and $\textbf{F}^\mathcal{Q}_i \in \mathbb{R} ^ { B_{\mathcal{Q}} \times L_i \times D'}$, where $i \in \left\{v,t\right\}$ refers the current visual or textual branch, $L_i$ indicate the branch's token length and $B_\mathcal{Q}$ denotes the numbers of query videos. CTPCM can be divided into {local prototype construction}, which focuses on the fusion of each token, and {global prototype construction}, which emphasizes fusion across the entire video, as shown in Fig.~\ref{fig:CTPCM}.

For {local prototype construction}, given reshaped $\textbf{P}^S_i \in \mathbb{R} ^ {  L_i \times 1 \times D'}$ and $\textbf{F}^\mathcal{Q}_i \in \mathbb{R} ^ {{ L_i \times  B_\mathcal{Q}}  \times D'}$, we first calculate the token-level cosine similarity between them to obtain $ \textbf{A}_{loc,i}^{S} \in \mathbb{R}^ { L_i \times 1  \times B_\mathcal{Q}}$. Then, a Softmax function is employed to normalize   $ \textbf{A}_{loc,i}^{S}$, generating similarity scores between query samples and the support prototype. Finally, the local prototype $\textbf{P}_{loc,i}^S$ is constructed through a weighted summation of all query samples based on their similarity scores:
 \begin{equation}\label{}
   \textbf{P}_{loc,i}^S = \mathrm{Softmax}( \mathrm{Cosine}(\textbf{P}_i^S,\textbf{F}^\mathcal{Q}_i )) \cdot   \textbf{F}^\mathcal{Q}_i
    \end{equation}
Note that for the textual branch, due to the varying input lengths of the ``Known" type prompts between the query set and support set, their ($\textbf{P}_t^\mathcal{S}$ and $\textbf{F}_t^\mathcal{Q}$) token lengths differ as well. Therefore, we pad zeros to the shorter ones to ensure their lengths are consistent.

For {global prototype construction}, we initially apply global average pooling to $\textbf{P}^S_i$ and  $\textbf{F}^\mathcal{Q}_i$ 
  along the token dimension to obtain the video summary feature  $\textbf{F}^\mathcal{Q}_{i\_{avg}} \in \mathbb{R} ^ { B_\mathcal{Q} \times D'}$ and prototype $\textbf{P}^S_{i\_avg} \in \mathbb{R} ^ { 1 \times D'} $. Next, we calculate the video-level cosine similarity between them, resulting in $\textbf{A}_{glb,i} ^S  \in \mathbb{R}^ {1  \times B_\mathcal{Q}} $, followed by a Softmax function for normalization.  {The global prototype $\textbf{P}_{glb,i}^S$ is constructed as:}
 \begin{equation}\label{}
   \textbf{P}_{glb,i}^S = \mathrm{Softmax}( \mathrm{Cosine}(\textbf{P}^S_{i\_avg}, \textbf{F}^\mathcal{Q}_{i\_{avg}} )) \cdot  \textbf{F}^\mathcal{Q}_{i\_{avg}}
    \end{equation}

{In Eq. 5 and Eq. 6, we consider the transductive setting, which is a standard operation that utilizes the overall information from the query set to guide the construction of the support prototype.} Finally, we combine the original prototype $\textbf{P}_i^S$ with the local $\textbf{P}_{loc,i}^S$ and global prototype $\textbf{P}_{glb,i}^S$:
 \begin{equation}\label{}
   \widetilde{\textbf{P}_i^S} = (\textbf{P}_i^S + \alpha \cdot \textbf{P}_{loc,i}^S + (1-\alpha) \cdot  \textbf{P}_{glb,i}^S) /2
    \end{equation}
where $\alpha \in \left[0,1 \right] $  is a learnable hyperparameter, $\widetilde{\textbf{P}_i^S} \in \left\{\widetilde{\textbf{P}_v^\mathcal{S}}, \widetilde{\textbf{P}_t^\mathcal{S}} \right\}$. 
Notably, {local and global prototype construction} each has its strengths for different token types. {The former (LPC) is well-suited for textual branch tokens, as these tokens exhibit strong semantic alignment between support and query samples—largely due to the standardized format of input prompts, output answers, and the consistent word order produced by Video-LLaVA. The latter (GPC) is more effective when meeting visual branches because visual tokens vary significantly across frames, with the same-position tokens often differing semantically, making token-level matching less useful for most tasks in FSAR.}

\subsection{Multimodal Prototype Matching Metric (MPMM)}
\label{method_MPMM}
With the enhanced support prototypes $\widetilde{\textbf{P}^\mathcal{S}_v}$, $ \widetilde{\textbf{P}^\mathcal{S}_t}$ and the enhanced query features   ${\textbf{F}^\mathcal{Q}_v}$, ${\textbf{F}^\mathcal{Q}_t}$, we design a novel multimodal metric that fully capitalizes on the strengths of dual branches for efficient matching. To achieve this, we drew inspiration from the Bi-MHM~\cite{wang2022hybrid}, which modifies the Hausdorff distance for FSAR. Bi-MHM is a symmetric function that automatically identifies the best correspondences between two videos.
However, this approach is intended for unimodal models that cannot fully utilize the advantages of Video-LLaVA's rich multimodal output.

In our metric, we focus on extracting the most discriminative information for video recognition from the dual branches and effectively combining them for optimal matching. Initially, we calculate the dual branch's token level distances followed by Hausdorff and concatenate the dual outputs together, given by: 
\begin{equation}
\begin{aligned}
\mathbf{d}_{s,h}^{k,n}
= \mathrm{Cat}\left(
  \min_{\mathbf{q}^{g}_{v,j}\in \mathbf{q}^g_v}
    \bigl\lVert \mathbf{s}^{k,n}_{v,h}-\mathbf{q}^{g}_{v,j}\bigr\rVert, \right. \\
\left. \min_{\mathbf{q}^{g}_{t,j}\in \mathbf{q}^g_t}
    \bigl\lVert \mathbf{s}^{k,n}_{t,h}-\mathbf{q}^{g}_{t,j}\bigr\rVert
\right)
\end{aligned}
\label{eq8}
\end{equation}
\vskip -0.1in
\begin{equation}
\begin{aligned}
\mathbf{d}_{q,j}^{g}
= \mathrm{Cat}\left(
  \min_{\mathbf{s}^{k,n}_{v,h}\in \mathbf{s}^{k,n}_{v}}
    \bigl\lVert \mathbf{q}^{g}_{v,j}-\mathbf{s}^{k,n}_{v,h}\bigr\rVert, \right. \\
\left.
  \min_{\mathbf{s}^{k,n}_{t,h}\in \mathbf{s}^{k,n}_{t}}
    \bigl\lVert \mathbf{q}^{g}_{t,j}-\mathbf{s}^{k,n}_{t,h}\bigr\rVert
\right)
\end{aligned}
\label{eq9}
\end{equation}
where the $n$-th support video prototype in the $k$ class  represents $\textbf{s}^{k,n}_v{\in\mathbb{R}}^{T \times D'}$ for the visual branch and $\textbf{s}^{k,n}_t{\in\mathbb{R}}^{L_t \times D'}$ for the textual branch. Under the same rules, the $g$-th query video feature indicates $\textbf{q}^g_v{\in\mathbb{R}}^{T \times D'}$ and $\textbf{q}^g_t{\in\mathbb{R}}^{L_t \times D'}$. $\textbf{d}_{s,h}^{k,n}$ represents the token distances from the 
$h$-th support token to all query tokens, while  $\textbf{d}_{q,j}^{g}$ represents the token distances from the 
$j$-th query token to all support tokens. 

We observe that the textual tokens are significantly longer than the visual tokens, and much of their semantic information in textual tokens may not directly relate to the main action.  Directly summing the token distances between all tokens in $\textbf{d}_{s,h}^{k,n}$ and $\textbf{d}_{q,j}^{g}$ as the video-level distance not only introduces some noise but also reduces the influence of visual tokens. Therefore, we sum the top $u$ largest distances between  $\textbf{d}_{s,h}^{k,n}$ and $\textbf{d}_{q,j}^{g}$ to adaptively select the most action-related information from both visual and textual tokens. Through this approach, we reduce the impact of action-irrelevant information on matching decision-making, and our Multimodal Prototype Matching Metric can be expressed as:
\begin{equation}
\begin{aligned}
\mathbf{D}_M
= \frac{1}{u}\left(
  \sum_{\mathbf{d}_{s,h}^{k,n}\in \mathbf{d}_s^{k,n}}
    \mathrm{Top}_u\!\left(\mathbf{d}_{s,h}^{k,n}\right)
\right. \\
\left.
  + \sum_{\mathbf{d}_{q,j}^{g}\in \mathbf{d}_q^{g}}
    \mathrm{Top}_u\!\left(\mathbf{d}_{q,j}^{g}\right)
\right)
\end{aligned}
\end{equation}
where $u$  is an adjustable hyperparameter. During training, the negative distance $-\textbf{D}_M$ is used as the logit. For inference, we classify by selecting the support prototype closest to the query.

\section{Experiments}
\subsection{Experimental Setup}
\subsubsection{Datasts} We evaluate FSAR-LLaVA on five widely used benchmarks, categorized into spatial-related and temporal-related datasets. The former includes Kinetics~\cite{carreira2017quo}, HMDB51~\cite{kuehne2011hmdb}, and UCF101~\cite{soomro2012ucf101}, which involve relatively simple actions, allowing the category of some videos to be determined with just a single keyframe. The latter includes SSv2-Full~\cite{goyal2017something} and SSv2-Small~\cite{goyal2017something}, which involve more complex actions, requiring the model to have strong temporal modeling capabilities to understand the video. Meanwhile, for Kinetics, SSv2-Small, and SSv2-Full, we use the splits provided by ~\cite{zhu2018compound, cao2020few, zhu2021closer}, where 100 classes are selected and divided into 64/12/24 action classes for training, validation, and testing, respectively. For UCF101 and HMDB51, we evaluate our method using the splits provided by ~\cite{zhang2020few}.

\subsubsection{Network Arcitectures} 
We choose Video-LLaVA-7B~\cite{lin2023video} as our knowledge base,  extracting multimodal features from 31-$st$ (out of 32) of its Multimodal Decoder. This choice is validated through ablation in Sec.~\ref{layers_selection}.
In MFM, we set $D'=256$ and configure all attention modules to use a single layer, balancing performance and efficiency. In CPTCM, $\alpha$ is initialized to 0.1 for “Unknown” prompts and 0.9 otherwise. In MPMM, $u$ is set to 50 for spatial-related datasets under 5-way 5-shot settings, and 10 in all other cases.  Hyperparameter ablations are detailed in Sec.~\ref{hyper1} to Sec.\ref{layers_selection}. To demonstrate the generality of our framework, we select the Qwen2-VL-7B~\cite{wang2024qwen2}, {Qwen3-VL-2B~\cite{bai2025qwen3vltechnicalreport}, and Qwen3-VL-8B~\cite{bai2025qwen3vltechnicalreport}} as our knowledge base.
{In this pipeline, we similarly utilize the penultimate-layer hidden features produced by the multimodal decoder of Qwen2-VL {or Qwen3-VL}, while keeping all other configurations consistent with the FSAR-LLaVA pipeline.}

\subsubsection{Training and Inference}  Following TSN~\cite{wang2016temporal}, we uniformly sample 8 frames ($T$=8) from each video as input. These frames are augmented during training using basic techniques such as random horizontal flipping, cropping, and color jitter, while only center cropping is applied during inference. In training, 50,000 episodes are randomly sampled from the SSv2 dataset, while 10,000 episodes are sampled from each of the other datasets. We employ the Adam optimizer with the multi-step scheduler for training. Inference results are reported as the average performance across 10,000 tasks randomly sampled from the test sets of all datasets. 
For the prompt input of FSAR-LLaVA, we can utilize the ``Known" prompts to guide Video-LLaVA toward more focused learning by using a flexible question prompt constructed from the provided labels of the support set. Alternatively, we can switch to the ``Unknown” mode, employing a fixed question template: ``\textit{What's the action of the video?}", to alleviate the situation of missing labels in the wild. A detailed discussion of prompts can be found in Sec.~\ref{discuss_prompt}.

\begin{table*}[t!]
\centering
\caption{
{Comparison under 5-way k-shot settings on five widely-used benchmarks including HMDB51, UCF101, Kinetics, SSv2-Small, and SSv2-Full.}
}
\footnotesize
\setlength{\tabcolsep}{1.6pt}
\begin{tabular}{ccccccccccccc}
\toprule
 & & & 
\multicolumn{2}{c}{\textbf{HMDB51}} & \multicolumn{2}{c}{\textbf{UCF101}}   & \multicolumn{2}{c}{\textbf{Kinetics}} & \multicolumn{2}{c}{\textbf{SSv2-Small}} & \multicolumn{2}{c}{\textbf{SSv2-Full}} \\
 & \multirow{-2}{*}{\textbf{Method}}&  \multirow{-2}{*}{\textbf{Pre-training}}  & \textbf{1-shot} & \textbf{5-shot} & \textbf{1-shot} & \textbf{5-shot} & \textbf{1-shot} & \textbf{5-shot}  & \textbf{1-shot} & \textbf{5-shot}& \textbf{1-shot} & \textbf{5-shot} \\ \midrule
 \multirow{9.5}*{\rotatebox{90}{\shortstack{Data-driven \\ Unimodal}}} 
&OTAM~\cite{cao2020few}& INet-RN50 & 54.5 & 68.0 & 79.9 & 88.9  & 73.0 & 85.8 &36.4 &48.0  & 42.8  & 52.3   \\
&TRX~\cite{perrett2021temporal} & INet-RN50 &54.9 & 75.6 & 81.0 & 96.1  & 65.1 & 85.9  & 36.0 & 56.7 & 42.0 & 64.6   \\
&STRM~\cite{thatipelli2022spatio}  & INet-RN50 & 57.6 & 77.3 & 82.7 & 96.9  & 65.1 & 86.7 & 37.1 & 55.3 & 43.1 & 68.1 \\
&HyRSM~\cite{wang2022hybrid} & INet-RN50 &60.3& 76.0 & 83.9 & 94.7 & 73.7 & 86.1 & 40.6 & 56.1 & 54.3 & 69.0 \\
&HCL~\cite{zheng2022few} & INet-RN50  & 59.1 & 76.3 & 82.5 & 93.9  & 73.7 & 85.8 & 38.7  & 55.4   & 47.3 & 64.9   \\
&MoLo (OTAM)~\cite{wang2023molo}& INet-RN50 & 59.8 & 76.1 & 85.4 & 95.1  & 73.8 & 85.1 & 41.9 & 56.2 & 55.0 & 69.6 \\
&GgHM ~\cite{xing2023boosting} & INet-RN50 & 61.2 & 76.9 & 85.2 & 96.3 & 74.9 & 87.4 & - & - & 54.5 & 69.2   \\
&Huang $et al.$~\cite{huang2024matching}& INet-RN50  & 61.6 & 77.5 & 74.9  &  92.5  & 74.0 & 86.9 & 42.6  & 61.8  & 52.3 & 67.1 \\
&TEAM~\cite{lee2025temporal}& INet-RN50  & 62.8 &  78.4 & 87.2  &  96.2  & 75.1 & 88.2 & -  & -  & - & - \\
\midrule
 \multirow{5.5}*{\rotatebox{90}{\shortstack{Data-driven \\ Multimodal}}} 

&CLIP-FSAR ~\cite{wang2023clip}  & CLIP-ViT-B & 77.1 & 87.7 & {97.0} & {99.1} & {94.8} & {95.4} & 54.6 & 61.8 & 62.1 & {72.1}
\\ 
&CLIP-CPM$^2$C~\cite{yang2023aim}  & CLIP-ViT-B & 75.9  & 88.0  &   95.0 & 98.6 &  91.0 & {95.5} & 
52.3 & 62.6 & 62.1& {72.8}\\ 
&MVP-shot~\cite{qu2024mvp} & CLIP-ViT-B   &  77.0&  88.1& 96.8 & 99.0 & 91.0 & 95.1& 55.4 & 62.0 & -& -\\
& EMP-Net~\cite{wu2024efficient} & CLIP-ViT-B   &  {76.8} & {85.8} & {94.3} & {98.2}& {89.1} & {93.5} & {57.1} & {65.7} & {63.1} & {73.0}\\
& Task-Adapter~\cite{cao2024task} & CLIP-ViT-B   &  {83.6} & {88.8} & {98.0} & {99.4}& {95.0} & {96.8} & {60.2} & {70.2} & {71.3} & {74.2}\\
\midrule
 \multirow{18}*{\rotatebox{90}{\shortstack{Knowledge-driven}}}
& CapFSAR~\cite{wang2023few}  & BLIP-ViT-B & 70.3 & 81.3   &93.1 &  97.7 & 83.5  &92.2  & 45.8  & 61.1 &54.0  & 70.1 \\  
& Video-LLaVA$^{\mathrm{Frozen\_Avg}}_{\mathrm{Unknown}}$~\cite{lin2023video} & Video-LLaVA-7B & 61.0 & 81.2 & 91.9 & 97.9 & 79.3 & 92.6 & 32.7 &47.6  & 31.2 & 45.2
\\
& Video-LLaVA$^{\mathrm{Frozen\_Avg}}_{\mathrm{Known}}$~\cite{lin2023video} & Video-LLaVA-7B & 71.5 &  81.9 & 95.8 & 98.1 & 91.1 & 93.1 & 33.7 &  48.5 & 31.9 & 45.5
\\
& Qwen2-VL$^{\mathrm{Frozen\_Avg}}_{\mathrm{Unknown}}$~\cite{wang2024qwen2} & Qwen2-VL-7B & 66.6  & 84.6  & 95.0 &98.2 &83.0  & 93.1 &36.1  &55.1   & 35.3 & 53.5 \\
& Qwen2-VL$^{\mathrm{Frozen\_Avg}}_{\mathrm{Known}}$~\cite{wang2024qwen2}& Qwen2-VL-7B & 68.7 & 86.9  & 96.3 & 98.7 & 86.4 & 93.9 & 38.7 & 56.3  & 36.6 & 55.6  \\

& Qwen3-VL$^{\mathrm{Frozen\_Avg}}_{\mathrm{Unknown}}$~\cite{bai2025qwen3vltechnicalreport} & Qwen3-VL-2B & 69.2  & 86.7  & 95.9 &98.5 &85.6  & 94.7 & 37.2  & 57.2   & 36.0 & 54.5 \\
& Qwen3-VL$^{\mathrm{Frozen\_Avg}}_{\mathrm{Known}}$~\cite{bai2025qwen3vltechnicalreport}& Qwen3-VL-2B & 71.2 & 87.3 & 96.0 & 98.7 & 86.3 & 94.8 & 38.8 & 57.8 & 36.2 & 54.6  \\

& Qwen3-VL$^{\mathrm{Frozen\_Avg}}_{\mathrm{Unknown}}$~\cite{bai2025qwen3vltechnicalreport} & Qwen3-VL-8B & 70.7  & 88.5  & 97.2 &98.8 &86.0  & 95.2 &39.0  &61.5   & 37.5 & 59.2 \\
& Qwen3-VL$^{\mathrm{Frozen\_Avg}}_{\mathrm{Known}}$~\cite{bai2025qwen3vltechnicalreport}& Qwen3-VL-8B & 72.9 & 88.9  & 97.3 & 98.9 & 86.9 & 95.3 & 39.8 & 62.2  & 37.6 & 59.4  \\

& \textbf{FSAR-LLaVA$_\mathrm{Unknown}$} & Video-LLaVA-7B & {84.3} & {92.7} & {98.1}  & 99.7 & 94.8 & {97.4} & 51.0& {71.1} &56.7 & {76.7}\\
& \textbf{FSAR-LLaVA$_\mathrm{Known}$} & Video-LLaVA-7B & \underline{87.8} & {93.3} & {99.0} & \underline{99.8} & \underline{95.3} & \underline{98.1} & {65.5}& {76.1} &{71.7} & {78.2} \\ 
& \textbf{FSAR-Qwen2$_\mathrm{Unknown}$} & Qwen2-VL-7B
 &84.0  & 91.6 &98.6  & 99.7 & 93.4 & 97.1 & 53.6 & 71.3 & 56.3  & 77.2  \\
& \textbf{FSAR-Qwen2$_\mathrm{Known}$} & Qwen2-VL-7B
 &  {86.9} & 92.5 & 98.8  & \underline{99.8} & 94.2 & {97.8}  & {66.8}  & {75.9} &  {74.4} &  {78.8}  \\ 

& \textbf{FSAR-Qwen3$_\mathrm{Unknown}$} & Qwen3-VL-2B
 &85.2  & 92.5 &98.7  & {99.7} & 93.5 & 97.2 & 54.2 & 71.7 & 56.6  & 77.1  \\
& \textbf{FSAR-Qwen3$_\mathrm{Known}$} & Qwen3-VL-2B
 &  \underline{87.8} & \underline{93.4} & \underline{99.1}  & \underline{99.8} & 94.4 & {97.9}  & \underline{67.1}  & \underline{76.3} &  \underline{74.8} &  \underline{79.0}  \\ 

 & \textbf{FSAR-Qwen3$_\mathrm{Unknown}$} & Qwen3-VL-8B
 &86.1  & 92.9 &98.6  & \underline{99.8} & 94.9 & 97.6 & 55.0 & 71.9 & 57.1  & 77.6  \\
& \textbf{FSAR-Qwen3$_\mathrm{Known}$} & Qwen3-VL-8B
 &  \textbf{88.4} & \textbf{93.9} & \textbf{99.3}  & \textbf{99.9} & \textbf{95.6} & \textbf{98.4}  & \textbf{68.4}  & \textbf{76.9} &  \textbf{75.2} &  \textbf{79.3}  \\ 
\bottomrule
        \end{tabular}
\label{tab:results}
\end{table*}
 

\subsection{Results}
\subsubsection{Quantitative Analysis}
To demonstrate the effectiveness and generality of our FSAR framework, we conduct experiments on two MLLMs: Video-LLaVA~\cite{lin2023video}, Qwen2-VL~\cite{wang2024qwen2}, {and Qwen3-VL~\cite{bai2025qwen3vltechnicalreport}}, and compare their performance with SOTA data-driven unimodal, multimodal, and knowledge-driven methods across five widely used datasets. We first introduce four baselines, Video-LLaVA$^{\mathrm{Frozen\_Avg}}_{\mathrm{Unknown}}$, Video-LLaVA$^{\mathrm{Frozen\_Avg}}_{\mathrm{Known}}$, Qwen2-VL$^{\mathrm{Frozen\_Avg}}_{\mathrm{Unknown}}$, Qwen2-VL$^{\mathrm{Frozen\_Avg}}_{\mathrm{Known}}$, {Qwen3-VL$^{\mathrm{Frozen\_Avg}}_{\mathrm{Unknown}}$, and Qwen3-VL$^{\mathrm{Frozen\_Avg}}_{\mathrm{Known}}$}, to highlight the advantages of leveraging the knowledge base. These baselines directly use the penultimate-layer hidden features from the frozen MLLMs, combined with simple average pooling for matching, under either “Unknown” or “Known” prompt settings. The only difference between the “Known” and “Unknown” settings is the type of prompt used.


The results are shown in Tab.~\ref{tab:results}, our FSAR-LLaVA and FSAR-Qwen2 achieve outstanding performance compared to previous methods. Based on the experimental results, we can draw the following conclusions: 1) Our framework is general and effective. Our FSAR-LLaVA$_\mathrm{Known}$, FSAR-Qwen2$_\mathrm{Known}$, and {FSAR-Qwen3$_\mathrm{Known}$} collectively dominate the SOTA performance across all settings, surpassing both data-driven and knowledge-driven methods.
In most common industrial application scenarios where no prior label information is available, our FSAR-LLaVA$_\mathrm{Unknown}$,  FSAR-Qwen2$_\mathrm{Unknown}$, {and  FSAR-Qwen3$_\mathrm{Unknown}$},  consistently deliver outstanding performance, even surpassing most data-driven multimodal methods utilizing label information. Compared to the strongest data-driven method Task-Adapter~\cite{cao2024task}, FSAR-LLaVA$_\mathrm{Unknown}$ outperforms it in most settings, particularly on HMDB51, where the 5-shot result shows a 3.9\%  improvement. Compared to CapFSAR~\cite{wang2023few} using BLIP~\cite{li2022blip}, our FSAR-LLaVA allows more flexible prompt inputs and delivers better recognition performance. 2) Introducing MLLM as a knowledge base in FSAR is crucial. Even the simplest frozen Video-LLaVA$^{\mathrm{Frozen\_Avg}}_{\mathrm{Unknown}}$, Qwen2-VL$^{\mathrm{Frozen\_Avg}}_{\mathrm{Unknown}}$, and  {Qwen3-VL$^{\mathrm{Frozen\_Avg}}_{\mathrm{Unknown}}$} without any label priors perform better on spatial-related datasets than previous carefully designed data-driven unimodal methods. 

\subsubsection{Qualitative Analysis}
\begin{figure*} [t!]
	\centering
	\includegraphics[width=\linewidth]{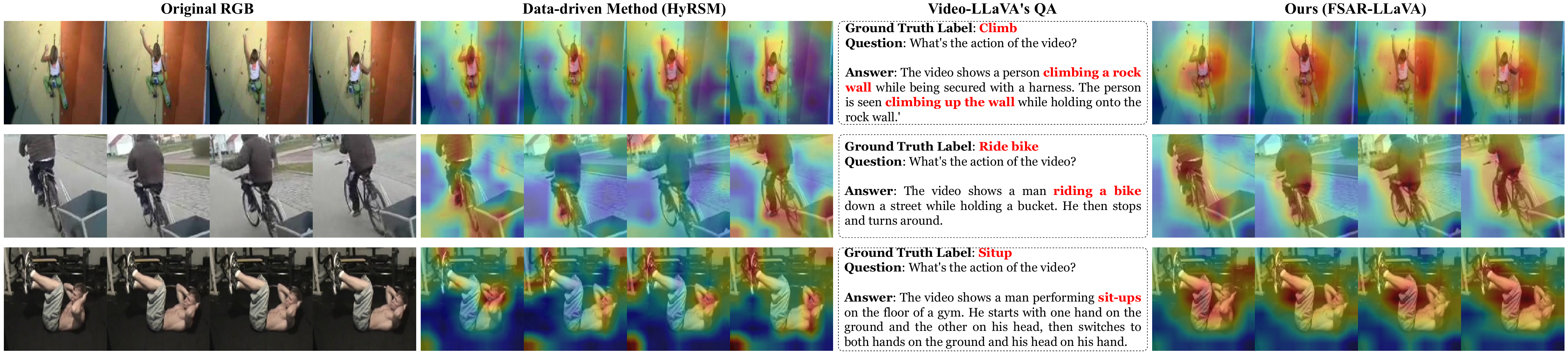}
    \caption{Visualization of the attention map for the data-driven method Hyrsm~\cite{wang2022hybrid} and our FSAR-LLaVA on HMDB51. We also provide the Video-LLaVA's QA results for these videos using ``Unknown" prompts.}
	\label{fig:vis}
\end{figure*}

To evaluate the disparity in video semantic understanding between data-driven and knowledge-driven methods without textual label priors, we conducted a visualization analysis of attention maps on HMDB51. This compared the data-driven method HyRSM~\cite{wang2022hybrid} with our FSAR-LLaVA using ``Unknown" prompts, as shown in Fig.~\ref{fig:vis}. The third column in the figure illustrates the QA results from directly applying Video-LLaVA to the videos, demonstrating its capability to accurately describe the video content. Comparing the second and fourth columns, we see that FSAR-LLaVA’s attention maps focus more on action-related objects while minimizing attention to irrelevant ones, highlighting the effectiveness of Video-LLaVA as our knowledge base in enhancing video semantic understanding.

\subsubsection{Effectiveness of Video-LLaVA in FSAR {and Failure Analysis}}
\label{discuss_prompt}
\begin{figure*} [t!]
	\centering
	\includegraphics[width=\linewidth]{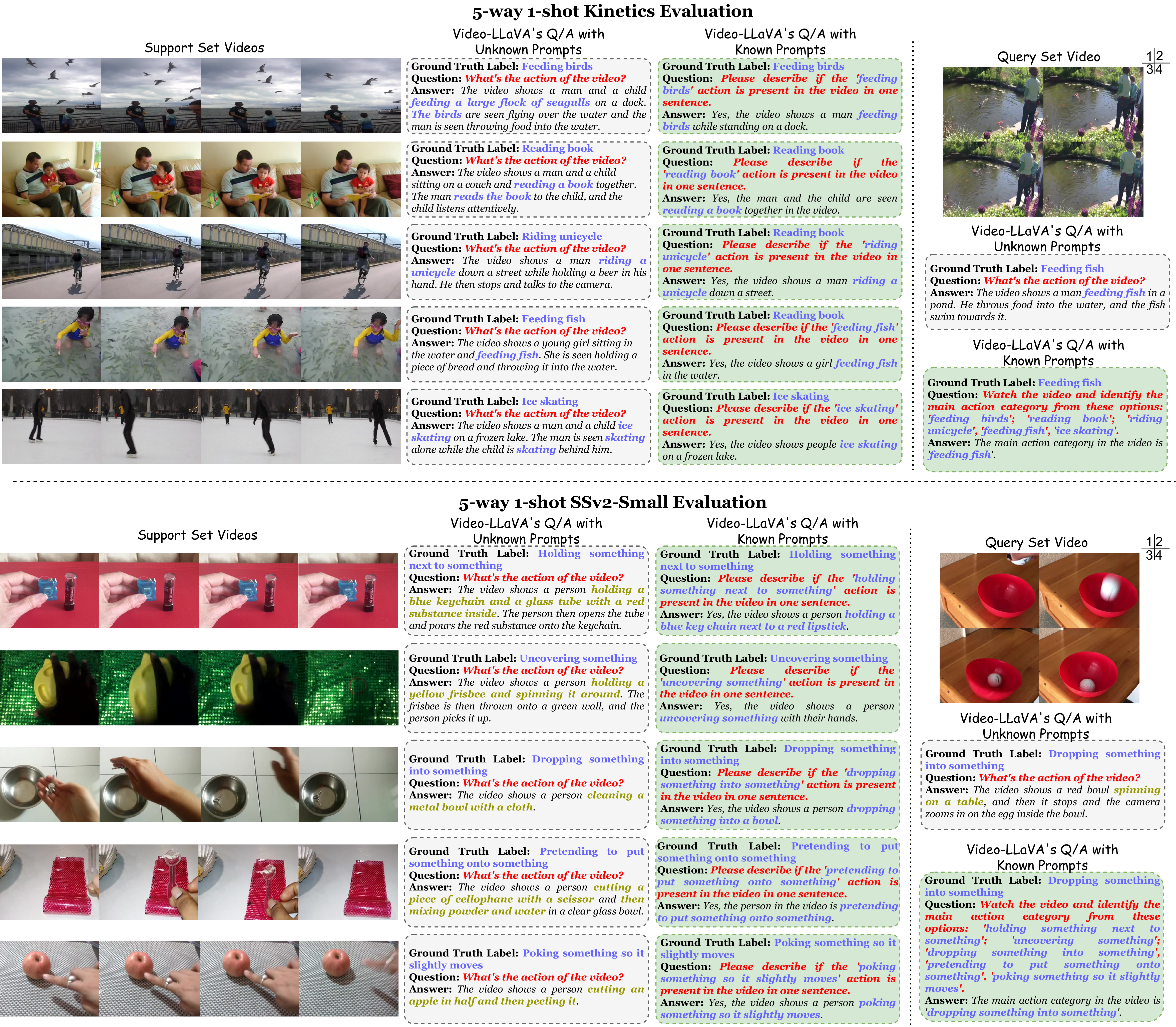}
	\caption{Direct qualitative analysis of the Video-LLaVA's output with different types of prompts. For clarity, we highlight words in \textcolor[HTML]{6666FF}{blue} to represent accurate action labels, in \textcolor{red}{red} to reveal the question posed in Video-LLaVA, and in  \textcolor[HTML]{999900}{green} to indicate inaccurate action descriptions.}
	\label{fig:llava}
\end{figure*}

We conduct experiments on two representative datasets to assess the effectiveness of Video-LLaVA~\cite{lin2023video} in few-shot action recognition (FSAR) and explore the performance differences between using ``Unknown" and ``Known" prompts. We apply Video-LLaVA directly to few-shot videos in a 5-way 1-shot setting. For ``Unknown" prompts, we do not provide any prior text labels, so all support and query videos are with the same question: ``\textit{What's the action of the video?}" For ``Known" prompts, we incorporate text priors into the video questions to make it easier for Video-LLaVA to understand the videos. Specifically, for the support set samples, we provide the question template: ``\textit{Please describe if the `\textbf{AC}' action is present in the video in one sentence}.'' where \textit{`\textbf{AC}'} represents the action category label of the current support sample. For the unseen query set, we restrict the response scope of Video-LLaVA, limiting it to select the query sample's category from a predefined set of categories within the current task, thereby reducing the difficulty of video comprehension. The question template is: \textit{``Watch the video and identify the main action category from these options: \textbf{`ACs'}." }where \textit{\textbf{`ACs'}} represents all action categories that appear in the support samples for the current task. The results are illustrated in Fig.~\ref{fig:llava}. For the spatial-related dataset Kinetics, Video-LLaVA accurately identifies action categories in all videos, even when using ``Unknown" prompts, demonstrating its strong comprehension of videos with simple actions. However, for the temporal-related dataset SSv2-Small, Video-LLaVA struggles with videos that involve complex, time-dependent actions, resulting in a mismatch using ``Unknown" prompts between the predicted action categories and the ground truth labels. By utilizing ``Known" prompts, we introduce textual priors related to the current video into the model, resolving this issue and enabling it to accurately identify action categories in unseen query videos. 
{This observation is also consistent with the
quantitative results reported in Tab.~\ref{tab:results}, where
\textbf{FSAR-LLaVA$_\mathrm{Unknown}$} (row 24) performs significantly worse than
\textbf{FSAR-LLaVA$_\mathrm{Known}$} (row 25) on SSv2-Small and SSv2-Full.
In summary, with the design of these two prompts for Video-LLaVA, our FSAR-LLaVA consistently outperforms the knowledge-driven CapFSAR~\cite{wang2023few} across all datasets, even when using the “Unknown” prompts. Moreover, when equipped with the “Known” prompts, FSAR-LLaVA achieves state-of-the-art performance on all datasets.}

\subsection{Ablation Study}
\subsubsection{Importance of Directly Employing the Hidden Features from MLLM}
{Tab.~\ref{pipe_diff} compares two pipelines under the 5-way 1-shot setting using both
``{Unknown}" and ``{Known}" prompts. The
``feature$\rightarrow$caption$\rightarrow$feature$\rightarrow$'' pipeline treats
Video-LLaVA as a black box by first generating video captions and then encoding them into
textual features, whereas the ``feature$\rightarrow$'' pipeline directly leverages
intermediate hidden features from Video-LLaVA. Both pipelines follow the same paradigm as
the visual branch of FSAR-LLaVA while excluding CTPCM, enabling a fair feature-level
comparison.
Under ``{Unknown}" prompts, the proposed ``feature$\rightarrow$'' pipeline consistently
outperforms the caption-based pipeline on spatial-related datasets, achieving gains of
2.4\% and 9.7\% on Kinetics and HMDB51, respectively. This highlights the advantage of
directly leveraging multimodal knowledge without going through the intermediate captioning
step, which may introduce information loss. In contrast, the performance gain on the
temporal-related dataset SSv2-Small remains marginal (only 0.4\%). As discussed in
Sec.~\ref{discuss_prompt}, this is mainly because Video-LLaVA itself struggles to accurately recognize
complex, time-dependent actions under ``{Unknown}" prompts. In this scenario, the
performance of both pipelines is heavily affected by the inaccurate video understanding of
the underlying MLLM, textual information plays a limited role, and the overall performance
is primarily dominated by visual cues. Consequently, regardless of how textual information
is utilized, the achievable performance gain on SSv2-Small remains limited.
To further illustrate the effectiveness of directly leveraging intermediate features when
task-relevant textual priors are available, we additionally report results under
``{Known}" prompts. As shown in Tab.~\ref{pipe_diff}, SSv2-Small achieves a much
more notable performance improvement (4.1\%) under the ``feature$\rightarrow$'' pipeline.
This result further confirms that the limited gains observed on SSv2-Small under
``{Unknown}" prompts stem from inaccurate textual representations rather than limitations
of the proposed pipeline design.}

\begin{table}[t!]
\caption{{Performance comparison with different pipelines.}}
\centering
 \setlength{\tabcolsep}{1.5pt}
\footnotesize
\begin{tabular}{lccc}
\toprule
\textbf{Pipeline} & \textbf{Kinetics} & \textbf{HMDB} & \textbf{SSv2-Small} \\ \midrule
\textit{\textbf{Using  ``Unknown"  Prompts}} & & &  \\ 
feature$\rightarrow$caption$\rightarrow$feature$\rightarrow$ & 85.7 & 69.8 & 49.5 \\
\textbf{feature$\rightarrow$} & \textbf{88.1} & \textbf{79.5} & \textbf{49.9} \\ \midrule
\textit{\textbf{Using  ``Known"  Prompts}} & & &  \\ 
feature$\rightarrow$caption$\rightarrow$feature$\rightarrow$ & 87.8 & 75.4 & 56.6 \\
\textbf{feature$\rightarrow$} & \textbf{91.1} & \textbf{83.8} & \textbf{60.7} \\ 
\bottomrule

\end{tabular}

\label{pipe_diff}
\end{table}

\subsubsection{Necessity of Multimodal Feature Decoupling}
\label{ness of decoup}
In Tab.~\ref{tab:Decoupling}, we conduct ablation experiments using ``Unknown'' prompts on four popular datasets to highlight the necessity of multimodal feature decoupling. We compare two training-free models: ``Avg.'', which applies global average pooling (GAP) to Video-LLaVA's penultimate-layer intermediate features followed by cosine similarity matching, and ``Dec-Avg.'', which first decouples these multimodal features into visual and textual branches, applies GAP and cosine similarity matching to each branch separately, and then sums the resulting scores.  Results show ``Dec-Avg.'' outperforms ``Avg.'' by 0.4\% to 2.8\% across all configurations, solely due to its decoupling of multimodal features. This result demonstrates the necessity of decoupling, as it reduces interference between different modality tokens during matching. Meanwhile, to further explore the preferences of different data on different branches of the model, we conduct experiments on two representative datasets in the 5-way 1-shot setting. The ``Visual Branch" involves final matching using only visual tokens, while the ``Textual Branch" operates under the same principle, utilizing only textual tokens. The results, shown in Tab.~\ref{tab:Branch}, indicate that the temporal-related dataset SSv2-Small favors the visual branch (``Unknown": 49.7 \% \textit{vs.} 50.6\%), while the spatial-related dataset  Kinetics prefers the textual branch (``Unknown": 94.8\%  \textit{vs.} 92.2\% ). Combining this with the text outputs of Video-LLaVA shown in Fig.~\ref{fig:llava}, we find that Video-LLaVA struggles with accurately understanding complex temporal videos. This leads to inaccurate semantic representations of textual tokens in the textual branch, resulting in poorer final matching performance on temporal-related datasets. Therefore, temporal-related datasets tend to favor using Video-LLaVA's visual branch for matching. However, Video-LLaVA behaves oppositely with spatial-related datasets. 



\begin{table}[t!]
\centering
  \caption{The Necessity of Multimodal Feature Decoupling. 
  ``Avg.'' and ``Dec-Avg.'' indicate the global average pooling and 
  global average pooling with decoupling the features.}
\label{tab:Decoupling}
\setlength{\tabcolsep}{2pt}
\footnotesize
\begin{tabular}{cccccccccc}
\toprule
 & 
\multicolumn{2}{c}{\textbf{Kinetics}} & \multicolumn{2}{c}{\textbf{HMDB51}}& \multicolumn{2}{c}{\textbf{UCF101}}& \multicolumn{2}{c}{\textbf{SSv2-Small}}   \\
\multirow{-2}{*}{\textbf{Method}}& \textbf{1s} & \textbf{5s} & \textbf{1s} & \textbf{5s} & \textbf{1s} & \textbf{5s} & \textbf{1s} & \textbf{5s}\\ \midrule
{Avg.} & 79.3 & 92.6 & 61.0 & 81.2 & 91.9 & 97.9  & 32.7& 47.6\\
\textbf{Dec-Avg.}  & \textbf{82.1}&  \textbf{93.5} & \textbf{63.3} & \textbf{82.7} & \textbf{93.2}& \textbf{98.3}  & \textbf{33.5} & \textbf{48.8}\\ \bottomrule
\end{tabular}
\end{table}

\begin{table}[t!]
\centering
\footnotesize
\setlength{\tabcolsep}{2pt}
\caption{Effect of different branches. ``Ukn." and ``Kn." refer to Unknown and Known.}
\label{tab:Branch}
\begin{tabular}{cccccccc}
\toprule
 & & 
\multicolumn{2}{c}{\textbf{Kinetics}} & \multicolumn{2}{c}{\textbf{SSv2-Small}}   \\
\multirow{-2}{*}{\textbf{Visual Branch}}& \multirow{-2}{*}{\textbf{Textual Branch}} & \textbf{Ukn.} & \textbf{Kn.}  & \textbf{Ukn.} & \textbf{Kn.} \\ \midrule
$\usym{2713}$ & $\usym{2717}$   & 92.2 &  93.3 & 50.6 & 62.2\\
$\usym{2717}$  & $\usym{2713}$  &94.3 & 94.5 & 49.7 & 63.9 \\ 
 $\usym{2713}$  &  $\usym{2713}$  &\textbf{94.8}& \textbf{95.3}& \textbf{51.0}& \textbf{65.5} \\ \bottomrule
\end{tabular}
\end{table}

\subsubsection{Effect of the Components in MFM}  To analyze the impact of the components in MFM, we conduct 5-way 1-shot experiments on Kinetics, SSv2-Small, and SSv2-Full, with results shown in Tab.~\ref{tab:MFM}.
The results indicate that the gradual introduction of the attention layer in our MFM yields significant performance gains on temporal-related datasets, while the improvements on spatial-related datasets (Kinetics) are comparatively smaller. The underlying cause of this phenomenon is that the frozen Video-LLaVA struggles with processing complex temporal information and cannot accurately understand temporal-related videos. As a remedy, our MFM enhances the features' representation through temporal and multimodal semantic relationship modeling, leading to performance enhancements on temporal-related datasets.

\begin{table}[t!]
\centering
\caption{Impact of the components in MFM. ``ST-SA", ``V-CA", and ``T-CA" indicate spatiotemporal self-attention, visual-lead cross-attention, and text-lead cross-attention, respectively.  ``Ukn." and ``Kn." refer to Unknown and Known.}
\label{tab:MFM}
\footnotesize
\setlength{\tabcolsep}{2pt}
\begin{tabular}{ccccccccc}
\toprule
 & & & 
\multicolumn{2}{c}{\textbf{Kinetics}} & \multicolumn{2}{c}{\textbf{SSv2-Small}}  & \multicolumn{2}{c}{\textbf{SSv2-Full}} \\
\multirow{-2}{*}{\textbf{ST-SA}}& \multirow{-2}{*}{\textbf{V-CA}}&  \multirow{-2}{*}{\textbf{T-CA}}  & \textbf{Ukn.} & \textbf{Kn.} & \textbf{Ukn.} & \textbf{Kn.} & \textbf{Ukn.} & \textbf{Kn.}  \\ \midrule
 $\usym{2717}$ &  $\usym{2717}$& $\usym{2717}$ &93.9 & 94.7  & 48.5 & 58.7& 51.4& 61.0  \\
$\usym{2713}$ & $\usym{2717}$ &  $\usym{2717}$   & 94.5&  95.0& 49.5&61.1  & 52.5& 63.7\\
 $\usym{2713}$  & $\usym{2713}$ &  $\usym{2717}$  &94.6 & 95.1 & 50.2 & 62.0  & 53.1   & 65.2 \\ 
 $\usym{2713}$  &  $\usym{2713}$ &  $\usym{2713}$  &\textbf{94.8}& \textbf{95.3}& \textbf{51.0} &\textbf{65.5} & \textbf{56.7} & \textbf{69.4} \\ \bottomrule
\end{tabular}
\end{table}

\subsubsection{Effect of the Components in CTPCM}
\label{effect_CTPCM}
To evaluate the impact of CTPCM, we perform ablation studies under the 5-way 1-shot setting on Kinetics, HMDB51, and SSv2. As shown in Tab.~\ref{tab:CTPCM}, spatial-related datasets tend to favor {local prototype construction}, whereas temporal-related datasets prefer {global prototype construction}. This is because Video-LLaVA excels in video understanding of spatial-related datasets, leading these videos to prefer the textual branch for matching. The tokens in the textual branch exhibit semantic consistency, making {local prototype construction} more effective. Conversely, temporal-related datasets tend to favor the visual branch, creating the opposite scenario.

\begin{table}[t!]
\centering
\caption{Impact of the components in CTPCM. ``LPC" and ``GPC" represent {local and global prototype construction}. ``Ukn." and ``Kn." refer to Unknown and Known.}
\label{tab:CTPCM}
\footnotesize
\setlength{\tabcolsep}{4pt}
\begin{tabular}{cccccccc}
\toprule
 & & 
\multicolumn{2}{c}{\textbf{Kinetics}} & \multicolumn{2}{c}{\textbf{HMDB51}}& \multicolumn{2}{c}{\textbf{SSv2-Small}}   \\
\multirow{-2}{*}{\textbf{LPC}}& \multirow{-2}{*}{\textbf{GPC}} & \textbf{Ukn.} & \textbf{Kn.} & \textbf{Ukn.} & \textbf{Kn.} & \textbf{Ukn.} & \textbf{Kn.} \\ \midrule
 $\usym{2717}$ &  $\usym{2717}$ &90.1 & 92.8 &79.0 & 83.5 &48.2 & 58.9  \\
$\usym{2713}$ & $\usym{2717}$   & 94.5&  95.0 & 83.7 & 86.9 &48.6&60.6 \\
$\usym{2717}$  & $\usym{2713}$  &92.9 & 93.9 & 82.8 &  86.1  &49.9 & 63.7 \\ 
 $\usym{2713}$  &  $\usym{2713}$  &\textbf{94.8}& \textbf{95.3}& \textbf{84.3}& \textbf{87.8}& \textbf{51.0} &\textbf{65.5} \\ \bottomrule
\end{tabular}
\end{table}

\subsubsection{Comparison of Different Matching Metrics}
{As shown in Tab.~\ref{tab:Metrics}, we compare the effectiveness of different matching metrics
under the 5-way 1-shot setting. Two baselines are considered: ``Hausdorff'', which employs the
Hausdorff distance as the matching metric, and ``Bi-MHM'', which applies Bi-MHM~\cite{wang2022hybrid}
to the visual and textual branches separately and sums the resulting matching scores. The
results demonstrate that the proposed MPMM, which is specifically designed for the dual-branch
architecture, consistently outperforms both baselines.
As described in Sec.~\ref{method_MPMM}, the key advantage of MPMM lies in its ability to jointly consider all
visual and textual tokens from both branches and dynamically select those most relevant to the
target action for matching. This advantage is particularly evident on the temporal-related
dataset SSv2-Small, where MPMM surpasses Bi-MHM by 2.1\% under ``{Unknown}" prompts and by
4.9\% under ``{Known}" prompts. The larger gains on SSv2-Small compared to Kinetics can be
attributed to two main factors. First, Kinetics is a relatively simpler dataset with higher
baseline accuracy, which leaves limited room for further improvement. Second, for
spatial-related datasets such as Kinetics, both visual and textual tokens are generally
informative for action recognition, thereby reducing the benefit of dynamic token selection.
By contrast, temporal-related datasets such as SSv2-Small are more challenging, where both
visual and textual branches are more likely to contain noisy or misleading tokens during
matching. In this scenario, dynamically selecting action-relevant tokens becomes significantly
more important, as it effectively suppresses the influence of noise tokens and leads to more
substantial performance gains.}

\begin{table}[t!]
\centering
\caption{Comparison of different matching metrics.}
\label{tab:Metrics}
\footnotesize
\setlength{\tabcolsep}{3pt}
\begin{tabular}{cccccccccc}
\toprule
 & 
\multicolumn{2}{c}{\textbf{Kinetics}} & \multicolumn{2}{c}{\textbf{SSv2-Small}}   \\
\multirow{-2}{*}{\textbf{Method}}& \textbf{Unknown} & \textbf{Known}  & \textbf{Unknown} & \textbf{Known}\\ \midrule
{Hausdorff} & 92.0 & 92.3  & 44.0 & 45.5\\
{Bi-MHM}  & 93.7 &  94.0 & 48.9 & 60.6\\
\textbf{Our MPMM} & \textbf{94.8} & \textbf{95.3}  & \textbf{51.0} & \textbf{65.5}   \\ \bottomrule
\end{tabular}
\end{table}


\subsubsection{Importance of Multimodal Knowledge} \label{importance_of_MLLM}
\begin{figure} [t!]
	\centering
	\includegraphics[width=\linewidth]{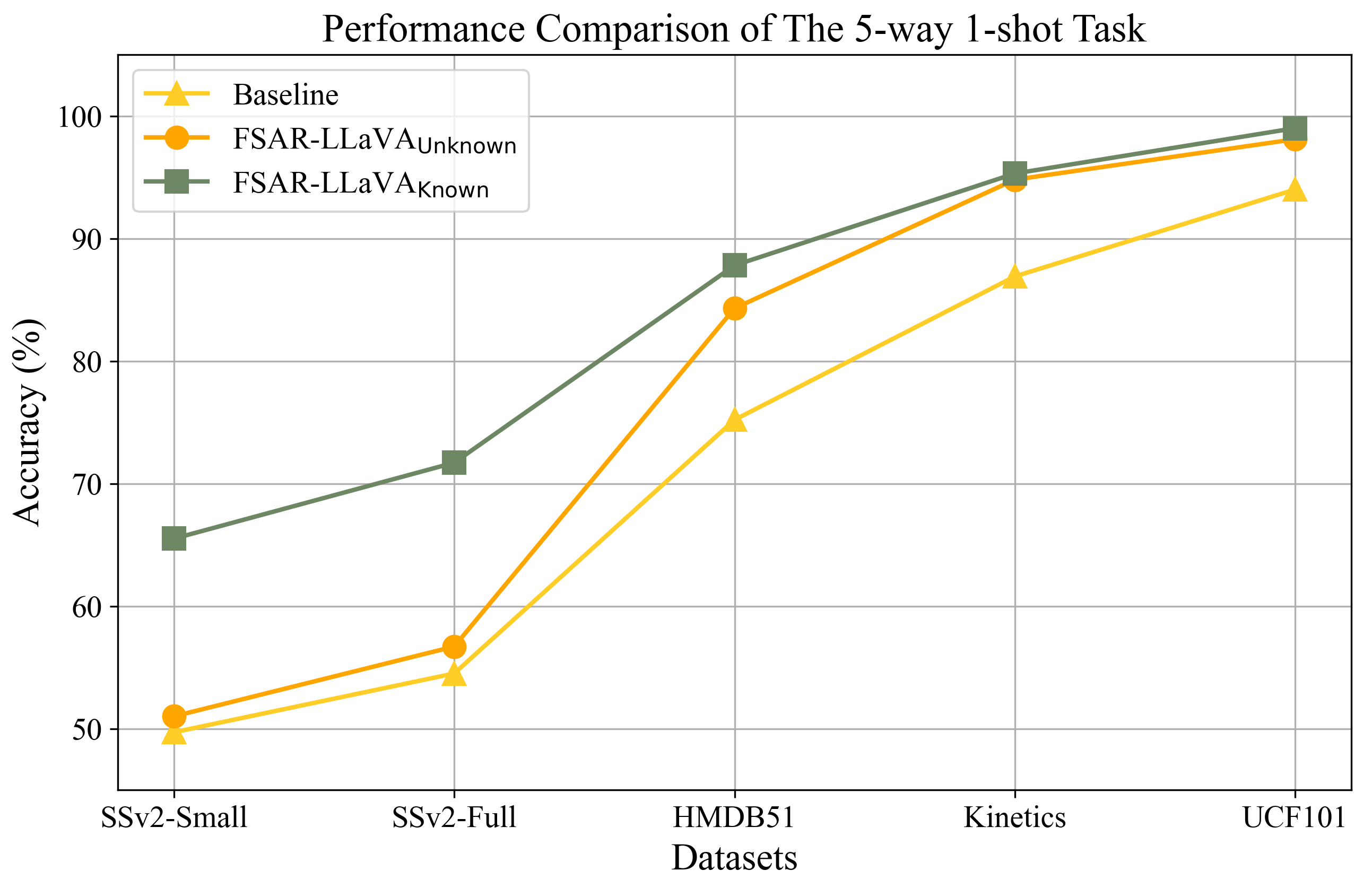}
	\caption{Performance comparison between our knowledge-driven FSAR-LLaVA and the data-driven baseline of the 5-way 1-shot task on five widely used datasets. All methods share the same vision encoder.}
	\label{fig:perf}
\end{figure}

To highlight the significance of multimodal knowledge in few-shot action recognition (FSAR), we conduct experiments on five widely used datasets under a 5-way 1-shot setting. We establish a baseline by substituting the multimodal features in our FSAR-LLaVA with the output of the video encoder from Video-LLaVA~\cite{lin2023video}  while retaining all the meticulously designed modules of our FSAR-LLaVA but excluding the textual branch and any textual information.
As shown in Fig.~\ref{fig:perf}, our FSAR-LLaVA$_\mathrm{Unknown}$ decisively outperforms the baseline model across all datasets, while our FSAR-LLaVA$_\mathrm{Known}$ leads by an even greater margin. The results demonstrate that multimodal knowledge is a crucial prior for FSAR.

\subsubsection{Discussion of Model Efficiency} \label{model_efficiency}
{The model efficiency results on a H20 GPU for the 5-way 1-shot task using
``{Unknown}" prompts are reported in Tab.~\ref{Tab:effi}. Here, ``Offline'' denotes the
setting where multimodal features extracted from the MLLM are pre-stored, while ``Online''
refers to real-time by feeding videos into the MLLM one by one.
Compared with data-driven methods such as OTAM~\cite{cao2020few},
HyRSM~\cite{wang2022hybrid}, and CLIP-FSAR~\cite{wang2023clip}, our FSAR-LLaVA (Offline)
achieves higher recognition accuracy with fewer trainable parameters, lower FLOPs, and
shorter time latency, demonstrating the efficiency of our framework.
To provide a clearer analysis of the computational bottlenecks, we explicitly decompose the
FLOPs into \emph{Backbone} (feature extraction networks, \textit{e.g.}, ResNet, ViT,
Video-LLaVA, and Qwen3-VL) and \emph{Module} (few-shot action recognition components,
including class-prototype enhancement, construction, and matching). This decomposition
offers a more intuitive understanding of the efficiency characteristics: the FLOPs
introduced by our modules are comparable to those of prior FSAR methods, while the dominant
computational overhead in the ``{Online}" setting mainly stems from the large MLLM
backbone, where videos are processed sequentially.
We also report results using Qwen3-VL-2B as the knowledge base of our framework
(Tab.~\ref{Tab:effi}, rows~6--7). Under this setting, the {online}
latency is approximately $2.5\times$ faster than that with Video-LLaVA, while achieving
strong recognition accuracy and remaining competitive with the larger Qwen3-VL-8B.
Moreover, compared with data-driven multimodal methods such as CLIP-FSAR, our framework (FSAR-Qwen3, row 7) achieves substantially higher accuracy with
a similar order of magnitude in latency per item.
These results demonstrate that adopting a lightweight MLLM as the knowledge base provides
a favorable trade-off between efficiency and performance within our framework.
}

\begin{table*}[t!]
\centering
\setlength\tabcolsep{5pt}
\footnotesize
\caption{{Complexity analysis for 5-way 1-shot HMDB51 evaluation. Our FSAR-LLaVA utilizes the ``Unknown" prompts here.
``Time" represents the training time cost of each 5-way 1-shot task.
``Acc" means the 5-way 1-shot accuracy on HMDB51.
All experiments are carried out on one Nvidia H20 GPU.}}
\begin{tabular}{cccccc}
\toprule
 & \textbf{Pre-Training} 
 & \textbf{Params} 
 & \textbf{FLOPs} 
 & \textbf{Time} 
 & \textbf{Acc} \\

 &  
 & \makecell{\textbf{Trainable}\textbf{/Full}} 
 & \makecell{\textbf{Backbone}\textbf{/Module}\textbf{/Total}} 
 &  
 &  \\
\midrule

OTAM~\cite{cao2020few}
& INet-RN50
& \makecell{23.5 M / 23.5 M}
&  32.8 G / 8.3 G / 41.1 G
& 0.48 s/item
& 54.5 \\

HyRSM~\cite{wang2022hybrid}
& INet-RN50
& \makecell{65.6 M / 65.6 M}
& 32.8 G / 8.5 G / 41.3 G
& 0.51 s/item
& 60.3 \\

CLIP-FSAR~\cite{wang2023clip}
& CLIP-ViT-B
& \makecell{90.2 M / 90.2 M}
& 140.8 G / 24.0 G / 164.8 G
& 1.28 s/item
& 77.1 \\

\textbf{FSAR-LLaVA (Offline)}
& Video-LLaVA-7B
& \makecell{9.3 M / 9.3 M}
& \makecell{- / 30.4 G / 30.4 G}
& 0.18 s/item
& 84.3 \\

\textbf{FSAR-LLaVA (Online)}
& Video-LLaVA-7B
& \makecell{9.3 M / 7 B}
& \makecell{30.19 T / 0.03 T / 30.22 T}
& 4.39 s/item
& 84.3 \\

\textbf{FSAR-Qwen3 (Offline)}
& Qwen3-VL-2B
& \makecell{7.2 M / 7.2 M}
& \makecell{- / 30.3G / 30.3 G}
& 0.17 s/item
& 85.2 \\

\textbf{FSAR-Qwen3 (Online)}
& Qwen3-VL-2B
& \makecell{7.2 M / 2 B}
& \makecell{8.63 T / 0.03 T / 8.66 T}
& 1.79 s/item
& 85.2 \\

\textbf{FSAR-Qwen3 (Offline)}
& Qwen3-VL-8B
& \makecell{9.3 M / 9.3 M}
& \makecell{- / 30.4 G / 30.4 G}
& 0.20 s/item
& 86.1 \\

\textbf{FSAR-Qwen3 (Online)}
& Qwen3-VL-8B
& \makecell{9.3 M / 8 B}
& \makecell{34.52 T / 0.03 T / 34.55 T}
& 4.45 s/item
& 86.1 \\

\bottomrule
\end{tabular}
\label{Tab:effi}
\end{table*}


\begin{table}[t!]
\setlength\tabcolsep{4pt}
\centering
\footnotesize
\caption{Effectiveness comparison with different mapping dimension $D'$.} 
{\begin{tabular}{cccc}
\toprule
  & \textbf{Trainable Params } &   \textbf{Kinetics}&  \textbf{SSv2-Small} \\ \midrule
$D'=128$ & 6.38M	      & 94.0 & 50.1  \\
    $D'=256$  	 & 9.26M  & 94.8   & 51.0 \\
 $D'=512$ & 16.52M     & 95.1 & 51.4\\
   $D'=1024$ & 37.02M    & 95.2 & 53.1\\ \bottomrule
\end{tabular}}
\label{D_disccuss}
\end{table}

\begin{table}[t!]
\centering
\caption{{Analysis of learnable parameter $\alpha$ in CTPCM.}}
\setlength{\tabcolsep}{3pt}
\footnotesize
\begin{tabular}{cccccccccc}
\toprule
 & \multicolumn{2}{c}{\textbf{Kinetics}} & \multicolumn{2}{c}{\textbf{SSv2-Small}}   \\
& \textbf{Unknown} & \textbf{Known}  & \textbf{Unknown} & \textbf{Known}\\ \midrule
$\alpha =0$ & 93.0 & 94.0 & 50.3 & 63.9\\
$\alpha =0.1$  & 93.3 &  94.2 & \textbf{51.0} & \textbf{65.5}\\
$\alpha =0.3$ & 93.9 & 94.4  & 49.9 & 63.7   \\  
$\alpha =0.5$ & 94.2 & 94.7  & 49.7 & 62.9   \\  
$\alpha =0.7$ & 94.5 & 94.9  & 49.4 & 62.3   \\ 
$\alpha =0.9$ & \textbf{94.8} & \underline{95.3}  & 49.2 & 61.7    \\  
$\alpha =1.0$ & \underline{94.7} & 95.1  & 48.9 & 60.9   \\ 
\textbf{Adaptive} & \underline{94.7} & \textbf{95.4}  & \underline{50.8} & \underline{65.3}  \\
\bottomrule
\end{tabular}
\label{tab:alpha}
\end{table}

\subsubsection{Discussion of Feature Mapping Dimension $D'$} 
\label{hyper1}
To determine the optimal feature mapping dimension, we perform experiments using ``Unknown" prompts in a 5-way 1-shot setting across two representative datasets. The results are shown in Tab.~\ref{D_disccuss}. The original Video-LLaVA's~\cite{lin2023video} intermediate feature dimension is 4096. We map this 4096 to $D'$, and all subsequent modules of our FSAR-LLaVA operate within the $D'$  dimension. As $D'$ increases, performance improves accordingly, as a higher feature dimension encapsulates more information. The performance improvement is particularly significant when $D'$ increases from 128 to 256, resulting in a 0.8\% performance increase on  Kinetics and a 0.9\% increase on  SSv2-Small. In contrast, the improvement is more limited when $D'$  increases from 256 to 512. Although the increase in $D'$ from 512 to 1024 leads to a significant improvement in SSv2-Small, it also substantially raises the number of trainable parameters. Ultimately, to balance performance and the number of trainable parameters, we set  $D'$ to 256, keeping the total trainable parameters of our FSAR-LLaVA within 10MB.

\subsubsection{{{Analysis of $\alpha$ in CTPCM and Exploration of Adaptive Gating Mechanism}}}
{
The parameter $\alpha$ is a learnable parameter that controls the balance between
local and global prototype construction in CTPCM. Specifically, when $\alpha$ approaches
0, global prototype construction (GPC) dominates, whereas values closer to 1 indicate that
local prototype construction (LPC) plays a more important role. 
To analyze the sensitivity of $\alpha$ and determine a suitable initial value for
different dataset types, we conduct experiments on the 5-way 1-shot task using two
representative datasets: the spatial-related dataset Kinetics and the temporal-related
dataset SSv2-Small. The results are reported in Tab.~\ref{tab:alpha}. 
Spatial-related datasets (\textit{e.g.}, Kinetics) improve smoothly with
larger $\alpha$, exhibiting a broad plateau over $\alpha\in[0.7,1.0]$, whereas temporal-related
datasets (\textit{e.g.}, SSv2-Small) perform best around $\alpha\approx0.1$ and remain stable within a
reasonable low-$\alpha$ range.}

{To further strengthen the framework's out-of-the-box generalization, we introduce a deterministic adaptive $\alpha$ assignment
applied consistently during both episodic training and inference. In this revised setting,
$\alpha$ is a non-trainable gating variable taking one of two fixed values $\{0.1,0.9\}$,
selected automatically per episode from matching statistics. This design is motivated by a
clear dataset-type trend: spatial-related datasets (HMDB51, UCF101, Kinetics) favor
textual-branch token matching (see Sec.~\ref{ness of decoup}) and, since textual tokens prefer LPC (see Sec.~\ref{CTPCM_intro}),
achieve the best results at $\alpha=0.9$; in contrast, temporal-related datasets
(SSv2-Small/Full) favor visual-branch matching and benefit more from GPC, thus preferring
$\alpha=0.1$. Moreover, when $\alpha$ was learnable in our earlier formulation, its optimized
value changed by less than $2\%$ relative to initialization, indicating limited benefit from
learning it and supporting the fixed $\{0.1,0.9\}$ choice.
Specifically, we apply the adaptive-$\alpha$ gating {after} MFM and {before} CTPCM.
MFM outputs the support prototypes $\mathbf{P}^{\mathcal{S}}_{v},\mathbf{P}^{\mathcal{S}}_{t}$
and the query features $\mathbf{F}^{\mathcal{Q}}_{v},\mathbf{F}^{\mathcal{Q}}_{t}$.
For a unified description, let $i\in\{v,t\}$ index the visual/textual branch. We denote the
corresponding token sequences as $\mathbf{p}^{k,n}_{i}\in\mathbb{R}^{L_i\times D'}$ for the
$n$-th support prototype of class $k$ and $\mathbf{f}^{g}_{i}\in\mathbb{R}^{L_i\times D'}$ for
the $g$-th query feature, where $\mathbf{p}^{k,n}_{i,h}$ is the $h$-th token of the support
prototype and $\mathbf{f}^{g}_{i,j}$ is the $j$-th token of the query feature (following the
subscript convention in Sec.~\ref{method_MPMM}, Eq.~(\ref{eq8})--(\ref{eq9})).
Here $L_v=T=8$ (8 frames), and we set $L_t=L_t'=8$ to suppress noise introduced by excessive
textual tokens.
Then, we define the branch-wise bi-directional Hausdorff-style distance:
\begin{equation}
\begin{aligned}
\mathbf{D}_i \;=\;&
\frac{1}{L_i}\sum_{\mathbf{p}^{k,n}_{i,h}\in \mathbf{p}^{k,n}_{i}}
\mathrm{Top}_{L_i}\left(\min_{\mathbf{f}^{g}_{i,j}\in \mathbf{f}^{g}_{i}}
\bigl\|\mathbf{p}^{k,n}_{i,h}-\mathbf{f}^{g}_{i,j}\bigr\|\right) +
\\
&
\frac{1}{L_i}\sum_{\mathbf{f}^{g}_{i,j}\in \mathbf{f}^{g}_{i}}
\mathrm{Top}_{L_i} \left(\min_{\mathbf{p}^{k,n}_{i,h}\in \mathbf{p}^{k,n}_{i}}
\bigl\|\mathbf{f}^{g}_{i,j}-\mathbf{p}^{k,n}_{i,h}\bigr\|\right)
\end{aligned}
\label{eq:Db_align_sec3_pf}
\end{equation}
We finally apply the hard gating (used in both training and inference):
\begin{equation}
\alpha=
\begin{cases}
0.1, & \text{if } \mathbf{D}_v < \mathbf{D}_t,\\
0.9, & \text{otherwise}.
\end{cases}
\label{eq:alpha_gate_align_sec3_pf}
\end{equation}
Intuitively, a smaller distance indicates more stable token-level matching. Thus, when the
visual branch is more stable (smaller $\mathbf{D}_{v}$), we favor LPC by setting a larger
$\alpha$; otherwise, we favor GPC by setting a smaller $\alpha$. This gating is
training-free and does not require learning additional parameters; it is applied identically
during training and inference. Moreover, it uses no query labels and introduces negligible
overhead. 
As shown in the last row of Tab.~\ref{tab:alpha}, with the proposed adaptive strategy, our
overall performance is on par with the best manually tuned (SOTA) setting across datasets.
Combined with our sensitivity analysis results, the proposed training-/inference-consistent
adaptive $\alpha$ selection preserves strong out-of-the-box generalization of our framework.}

\subsubsection{Analysis of adjustable parameter $u$ in MPMM}
$u$ is an integer parameter that regulates the amount of information involved in the final matching process within MPMM. To investigate the influence of 
$u$ on our model's performance across various datasets, we conduct experiments on  Kinetics, HMDB51, and SSv2-Small  using ``Unknown" prompts. As shown in Tab.~\ref{tab:u}, we find that for spatial-related datasets in 5-shot tasks, the optimal value of $u$ is 50, whereas in other scenarios, the optimal value of 
$u$ is 10. Typically, the combined total of visual and textual tokens exceeds 80, so when $u=50$, it already captures most of the information from both branches. As for the temporal-related dataset SSv2-Small, which primarily relies on visual tokens for matching, the performance is best when $u=10$, as there are only 8 visual tokens for 8 frames per video. For spatial-related datasets that primarily rely on textual tokens for matching, fewer support samples in a class tend to include more information irrelevant to the action subject within the tokens, while with more support samples, this issue diminishes. As a result, 
$u=10$ works best for 1-shot tasks, while 
$u=50$ is optimal for 5-shot tasks.
{Nevertheless, even with a unified setting of $u=10$ across all datasets, the overall
performance remains competitive with state-of-the-art settings, indicating that $u=10$ serves as a robust default choice for the framework.
Since a smaller $u$ focuses matching on the most discriminative information, we adopt $u=10$
as a default out-of-the-box choice to suppress redundant noise and preserve strong
performance on unseen datasets.}

\begin{table}[t!]
\centering
\caption{Analysis of adjustable parameter $u$ in MPMM.}
\setlength{\tabcolsep}{3pt}
\footnotesize
\begin{tabular}{ccccccc}
\toprule
 & \multicolumn{2}{c}{\textbf{Kinetics}} & \multicolumn{2}{c}{\textbf{HMDB51}}& \multicolumn{2}{c}{\textbf{SSv2-Small}}   \\
 & \textbf{1-shot} & \textbf{5-shot} & \textbf{1-shot} & \textbf{5-shot} & \textbf{1-shot} & \textbf{5-shot}  \\ \midrule
$u=10$  &  \textbf{94.8} & 97.1 &\textbf{84.3} & 92.4 & \textbf{51.0} & \textbf{71.1} \\
$u= 20$   & 94.7 &  97.2 & 84.2 & 92.4 &50.6&70.2 \\
$u= 30 $  &94.7 & 97.2 & 84.2 &  92.5 &50.1 & 69.3 \\ 
$u= 40$   &94.5 & 97.3 & 84.1 &  92.6 &49.9 & 68.6 \\ 
$u= 50$ &94.4 & \textbf{97.4} & 84.0 &  \textbf{92.7}  &49.7 & 68.3 \\ 
$u= 60$    & 94.3  &  \textbf{97.4} &  84.0 & 92.6  &  49.6 &  68.1\\ \bottomrule
\end{tabular}
\label{tab:u}
\end{table}

\subsubsection{Selection of the hidden layer features from Video-LLaVA}
\label{layers_selection}
In Tab.~\ref{hidden_layer}, we discuss the performance of different hidden layer features on five widely used datasets with ``Unknown" prompts under the 5-way 1-shot setting. To achieve a clear conclusion while avoiding any interference, we directly input the hidden features into our training-free MFMM to obtain matching scores. From the results, the 30-$th$ layer features achieve the best performance on HMDB51 and UCF101, the 31-$st$ layer features perform best on Kinetics, the 18th layer features excel on SSv2-Small, and the 22-$nd$, 23-$rd$, and 24-$th$ layer features show the best results on SSv2-Full. From the average results of the five datasets, the 31$st$ layer features perform the best, followed by 22$nd$ layer features. Based on the above results, we ultimately choose the 31-$st$ layer feature as our multimodal knowledge, given its relatively strong performance across all datasets. We also observe that the performance of the last layer is significantly worse compared to the second-to-last layer, likely due to the better semantic representation of the latter. As a result, when using Qwen2-VL as the knowledge base, we opt for the features from the second-to-last layer.

\begin{table}[t!]
\centering
\caption{Performance comparison of the hidden layer features from Video-LLaVA. The \textbf{boldfacen} and \underline{underline font} indicate the highest and the second highest results. Note that ``Avg." denotes Average.} 
\footnotesize
\setlength\tabcolsep{1pt}
\begin{tabular}{ccccccc}
\toprule
  &   \textbf{Kinetics}&   \textbf{HMDB}& \textbf{UCF}& \textbf{SSv2-Small} &  \textbf{SSv2-Full} &  \textbf{Avg.}  \\ \midrule
$l=0$ & 76.8 & 58.7 & 91.1 & 29.5 & 30.3 & 57.3 \\
$l=1$ & 76.7 & 58.4 & 91.1 & 29.2 & 30.1 & 57.1 \\
$l=2$ & 76.6 & 58.4 & 91.0 & 28.9 & 30.0 & 57.0 \\
$l=3$ & 76.4 & 58.7 & 91.4 & 29.2 & 29.8 & 57.1 \\
$l=4$ & 76.5 & 59.0 & 91.2 & 29.1 & 29.6 & 57.1 \\
$l=5$ & 77.6 & 59.4 & 91.8 & 29.5 & 29.2 & 57.5 \\
$l=6$ & 78.7 & 60.5 & 92.1 & 29.7 & 30.2 & 58.2 \\
$l=7$ & 82.1 & 64.0 & 93.1 & 31.2 & 31.0 & 60.3 \\
$l=8$ & 82.4 & 65.0 & 93.4 & 31.4 & 31.3 & 60.7 \\
$l=9$ & 82.3 & 65.1 & 93.4 & 32.1 & 31.6 & 60.9 \\
$l=10$ & 82.6 & 66.8 & 93.7 & 32.9 & 32.5 & 61.7 \\
$l=11$ & 82.5 & 66.8 & 93.6 & 33.3 & 32.6 & 61.8 \\
$l=12$ & 81.4 & 67.4 & 93.5 & 33.3 & 33.5 & 61.8 \\
$l=13$ & 80.5 & 66.9 & 93.5 & 34.0 & 33.9 & 61.8 \\
$l=14$ & 79.7 & 66.6 & 93.3 & 34.1 & 34.7 & 61.7 \\
$l=15$ & 80.9 & 68.4 & 94.0 & 35.2 & 35.8 & 62.9 \\
$l=16$ & 80.9 & 68.9 & 94.1 & 36.1 & 36.5 & 63.3 \\
$l=17$ & 82.6 & 68.3 & 93.9 & 36.4 & \underline{36.9} & 63.6 \\
$l=18$ & 84.1 & 69.9 & 94.7 & \textbf{37.0 } & 36.8 & 64.5 \\
$l=19$ & 84.2 & 69.6 & 95.3 & 36.7 & 36.7 & 64.5 \\
$l=20$ & 84.5 & 68.3 & 95.4 & 36.3 & 36.6 & 64.2 \\
$l=21$ & 85.1 & 69.3 & 95.5 & 36.4 & 36.8 & 64.6 \\
$l=22$ & 86.6 & 69.6 & 95.9 & \underline{36.8} & \textbf{37.2} & \underline{65.2} \\
$l=23$ & 86.5 & 69.5 & 95.8 & 36.6 & \textbf{37.2} & 65.1 \\
$l=24$ & 86.7 & 69.4 & 95.9 & 36.1 & \textbf{37.2} & 65.1 \\
$l=25$ & 87.0 & 69.0 & 96.1 & 35.5 & 36.3 & 64.8 \\
$l=26$ & 87.0 & 68.7 & 96.1 & 35.5 & 35.9 & 64.6 \\
$l=27$ & 87.0 & 68.9 & 96.2 & 35.5 & 36.2 & 64.8 \\
$l=28$ & 86.7 & 69.0 & 96.3 & 35.7 & 36.4 & 64.8 \\
$l=29$ & 87.1 & 69.5 & 96.4 & 35.6 & 36.1 & 64.9 \\
$l=30$ & \textbf{87.7} & \underline{69.7} & \textbf{96.6} & 35.4 & 36.3 & {65.1} \\
$l=31$ & \underline{87.4} & \textbf{70.4} & \underline{96.4} & 35.9 & 36.7 & \textbf{65.4} \\
$l=32$ & 84.7 & 68.3 & 94.8 & 33.9 & 34.6 & 63.3 \\ \bottomrule
\end{tabular}
\label{hidden_layer}
\end{table}

\begin{table*}[t!]
\centering
\caption{{Effect of different ``Unknown'' prompts on temporal-related datasets using different MLLMs within our framework.}}
\setlength{\tabcolsep}{3pt}
\footnotesize
\begin{tabular}{ccccccccc}
\toprule
\textbf{Prompt} & \textbf{Pre-training} & \multicolumn{2}{c}{\textbf{SSv2-Small}} & \multicolumn{2}{c}{\textbf{SSv2-Full}}   \\
& & \textbf{1-shot} & \textbf{5-shot}  & \textbf{1-shot} & \textbf{5-shot}\\ \midrule
\textit{``What's the action of the video?"} & Video-LLaVA-7B & 51.0 & 71.1 & 56.7 & 76.7\\
\textit{``Describe the sequence of actions happening in the video."} & Video-LLaVA-7B & 51.3 & 71.0 & 56.5 & 76.4\\
\textit{``How does the interaction between the person and the object evolve?"} & Video-LLaVA-7B & \textbf{52.0} & \textbf{71.6} & \textbf{57.1} & \textbf{77.2}\\ \midrule
\textit{``What's the action of the video?"} & Qwen3-VL-2B & 54.2 & 71.7  & 56.6 & 77.1   \\ 
\textit{``Describe the sequence of actions happening in the video."} & Qwen3-VL-2B & 54.1 & 71.5  & 56.7 & 77.0   \\ 
\textit{``How does the interaction between the person and the object evolve?"} & Qwen3-VL-2B & \textbf{54.9} & \textbf{72.3}  & \textbf{57.2} & \textbf{77.8} \\ \bottomrule
\end{tabular}
\label{tab:prompt_unknown}
\end{table*}

\subsubsection{{Effect of different “Unknown” prompts on temporal-related datasets}}
\label{different_prompt_effect}
\noindent {While our initial {``Unknown''} setting adopts a simple generic prompt
(\textit{``What’s the action of the video?''}), we note that {``Unknown''} prompts are not
limited to this formulation and still allow for meaningful design choices without introducing
explicit category labels. Along this line, we explore two alternative directions for improving
prompt design under the {``Unknown''} setting.
The first direction focuses on temporal description, such as
\emph{``Describe the sequence of actions happening in the video.''}, which aims to encourage
MLLMs to attend to temporal evolution. However, as shown in Tab.~\ref{tab:prompt_unknown} (rows 2 and 5), this
temporal-oriented prompt leads to nearly identical performance compared to the baseline
generic prompt across both FSAR-LLaVA and FSAR-Qwen3, suggesting limited improvement in complex
temporal reasoning under purely descriptive guidance.
The second direction adopts an interaction-focused formulation. As shown in Fig.~\ref{fig:llava}, action categories in SSv2 typically involve
a human performing operations on one or more objects. Motivated by this observation, we design
interaction-focused prompts such as
\emph{``What is the person doing with the object in the video?''}, which explicitly encourage
the model to reason about motion and object interaction. As reported in Tab.~\ref{tab:prompt_unknown} (rows 3 and 6),
this prompt consistently yields more noticeable gains (generally exceeding $0.5\%$ across
settings) compared to the baseline, indicating that adding mild structural constraints can
better activate motion and interaction reasoning even without explicit label supervision.
These results suggest that while generic {``Unknown''} prompts are insufficient for
eliciting strong temporal reasoning from frozen MLLM backbones, carefully designed and
label-free prompts, especially those emphasizing interaction semantics, can partially
improve the utilization of multimodal knowledge. }

\section{Conclusion}
We propose FSAR-LLaVA, a novel framework that directly exploits hidden features from the multimodal decoder of MLLMs for FSAR, with Video-LLaVA as the knowledge base. To fully leverage its multimodal and world knowledge capabilities, we introduce a Multimodal Feature-Enhanced Module, a Composite Task-Oriented Prototype Construction Module, and a Multimodal Matching Metric. Experiments on five benchmark datasets show that FSAR-LLaVA delivers strong performance with minimal trainable parameters.

\section{{Limitations}}
\noindent
{Our current exploration of prompt design remains limited, as this
paper only considers fixed prompt templates under the ``{Known}" and ``{Unknown}"
settings. As evidenced by the failure analysis in Sec.~\ref{discuss_prompt} and Fig.~\ref{fig:llava} of the manuscript, under the ``{Unknown}" setting, Video-LLaVA can frequently produce incorrect responses on temporal-related datasets, indicating that it is difficult to reliably recognize complex and time-dependent actions when no task-specific textual prior is provided. Meanwhile, as discussed in Sec.~\ref{different_prompt_effect}, even within the same ``{Unknown}" setting, adopting more guided yet label-agnostic fixed prompts can consistently affect and improve few-shot action recognition (FSAR) performance, suggesting that prompt formulation plays a non-trivial role in eliciting temporal reasoning.
Therefore, we believe there is considerable potential to further investigate more flexible prompting schemes, such as learnable continuous prompts (\textit{e.g.}, soft/prefix prompts with minimal additional parameters) or task-adaptive prompt selection/generation, to better exploit the video understanding capabilities of MLLMs (\textit{e.g.}, Video-LLaVA and Qwen2/3-VL) for FSAR. 
Importantly, such directions are expected to benefit FSAR in a unified manner, improving robustness across both ``{Known}'' and ``{Unknown}'' settings, irrespective of whether support-set textual labels are available.}

\noindent \textbf{Data Availability.} All experiments are conducted on publicly available datasets. Specifically, the SSV2 dataset~\cite{goyal2017something} is available at \href{https://developer.qualcomm.com/software/ai-datasets/something-something}{https://developer.qualcomm.com/software/ai-datasets/something-something}.  
The Kinetics dataset~\cite{carreira2017quo} can be found at \href{https://www.deepmind.com/open-source/kinetics}{https://www.deepmind.com/open-source/kinetics}.  
The HMDB-51 dataset~\cite{kuehne2011hmdb} is assessed at  \href{https://serre-lab.clps.brown.edu/resource/hmdb-a-large-human-motion-database}{https://serre-lab.clps.brown.edu/resource/hmdb-a-large-human-motion-database}.  
The UCF-101 dataset~\cite{soomro2012ucf101} is hosted at \href{https://www.crcv.ucf.edu/data/UCF101.php}{https://www.crcv.ucf.edu/data/UCF101.php}.

\bibliography{sn-bibliography}

\end{document}